\documentclass{article}

\usepackage{microtype}
\usepackage{graphicx}
\usepackage{subcaption}
\usepackage{booktabs} 
\usepackage{xcolor}
\definecolor{promptbg}{RGB}{245,245,245}
\usepackage[most]{tcolorbox}
\usepackage{pifont} 
\tcbuselibrary{breakable, skins}
\usepackage{caption}
\usepackage{hyperref}

\usepackage[preprint]{icml2026}

\usepackage{amsmath}
\usepackage{amssymb}
\usepackage{mathtools}
\usepackage{amsthm}
\usepackage{amsmath,amsfonts}
\usepackage{algorithmic}
\usepackage{array}

\usepackage{textcomp}
\usepackage{stfloats}
\usepackage{url}
\usepackage{verbatim}
\usepackage{graphicx}
\usepackage{booktabs}
\usepackage{amssymb}
\usepackage{mathtools}
\usepackage{amsthm}
\usepackage{xcolor}
\usepackage{multirow} 
\usepackage{changepage}
\usepackage{xcolor}
\usepackage{subcaption}
\usepackage{hyperref}
\usepackage{colortbl}
\usepackage[table]{xcolor}
\usepackage{enumitem}

\usepackage{titlesec}

\titlespacing*{\paragraph}{0pt}{0pt}{0.25em}

\usepackage[capitalize,noabbrev]{cleveref}

\theoremstyle{plain}

\theoremstyle{definition}

\theoremstyle{remark}

\usepackage[textsize=tiny]{todonotes}

\icmltitlerunning{Recycling Failures: Salvaging Exploration in RLVR via Fine-Grained Off-Policy Guidance}

\begin{document}

\twocolumn[
  \icmltitle{Recycling Failures: Salvaging Exploration in RLVR via Fine-Grained Off-Policy Guidance}




%
    \begin{icmlauthorlist}
      \icmlauthor{Yanwei Ren}{buaa,hiii}
      \icmlauthor{Haotian Zhang}{buaa,hiii}
      \icmlauthor{Likang Xiao}{hiii}
      \icmlauthor{Xikai Zhang}{hiii}
      \icmlauthor{Jiaxing Huang}{polyu}
      \icmlauthor{Jiayan Qiu}{leicester}
      \icmlauthor{Baosheng Yu}{ntu}
      \icmlauthor{Quan Chen}{kuaishou}
      \icmlauthor{Liu Liu}{buaa,hiii}
    \end{icmlauthorlist}
    
    \icmlaffiliation{buaa}{School of Artificial Intelligence, Beihang University, Beijing, China}
    \icmlaffiliation{hiii}{Hangzhou International Innovation Institute, Beihang University, Hangzhou, China}
    \icmlaffiliation{leicester}{University of Leicester, Leicester, United Kingdom}
    \icmlaffiliation{ntu}{Nanyang Technological University, Singapore}
    \icmlaffiliation{polyu}{The Hong Kong Polytechnic University, Hong Kong SAR, China}
    \icmlaffiliation{kuaishou}{Kuaishou Technology, Beijing, China}
    
    \icmlcorrespondingauthor{Liu Liu}{liuliubh@buaa.edu.cn}



  \icmlkeywords{Machine Learning, ICML}

  \vskip 0.3in
]



\printAffiliationsAndNotice{}  

\begin{abstract}

Reinforcement Learning from Verifiable Rewards (RLVR) has emerged as a powerful paradigm for enhancing the complex reasoning capabilities of Large Reasoning Models. However, standard outcome-based supervision suffers from a critical limitation that penalizes trajectories that are largely correct but fail due to several missteps as heavily as completely erroneous ones. This coarse feedback signal causes the model to discard valuable largely correct rollouts, leading to a degradation in rollout diversity that prematurely narrows the exploration space. Process Reward Models have demonstrated efficacy in providing reliable step-wise verification for test-time scaling, naively integrating these signals into RLVR as dense rewards proves ineffective.Prior methods attempt to introduce off-policy guided whole-trajectory replacement that often outside the policy model's distribution, but still fail to utilize the largely correct rollouts generated by the model itself and thus do not effectively mitigate the narrowing of the exploration space. To address these issues, we propose \textbf{SCOPE} (\textbf{S}tep-wise \textbf{C}orrection for \textbf{O}n-\textbf{P}olicy \textbf{E}xploration), a novel framework that utilizes Process Reward Models to pinpoint the first erroneous step in suboptimal rollouts and applies fine-grained, step-wise off-policy rectification. By applying precise refinement on partially correct rollout, our method effectively salvages partially correct trajectories and increases diversity score by 13.5\%, thereby sustaining a broad exploration space. Extensive experiments demonstrate that our approach establishes new state-of-the-art results, achieving an average accuracy of 46.6\% on math reasoning and exhibiting robust generalization with 53.4\% accuracy on out-of-distribution reasoning tasks.

\end{abstract}
\section{Introduction}


The advent of Reinforcement Learning from Verifiable Rewards (RLVR) has established a powerful paradigm for enhancing the complex reasoning capabilities of Large Reasoning Models (LRMs). By optimizing models against ground-truth outcomes, recent milestones such as DeepSeek-R1~\cite{guo2025deepseek} and DeepSeekMath~\cite{shao2024deepseekmath} have demonstrated that reinforcement learning can incentivize emergent reasoning behaviors that supervised fine-tuning alone fails to elicit. A representative example is Group Relative Policy Optimization (GRPO)~\cite{shao2024deepseekmath}, which eliminates the need for a value network by normalizing rewards within a group of sampled outputs. Despite these advances, standard RLVR methods fundamentally rely on sparse, outcome-based supervision, which assigns credit only to fully correct solutions. As a consequence, trajectories that are largely correct but contain several intermediate errors receive the same zero reward as completely incorrect attempts. This indiscriminate penalization forces the learning process to discard a vast number of informative, near-correct rollouts, leading to a critical \textbf{degradation in diversity}. This loss of diversity inevitably causes a premature narrowing of the exploration space and \textbf{loss of sample efficiency}, preventing the model from discovering and reinforcing complex reasoning paths that lie just beyond its current capabilities.


To mitigate this sample inefficiency and exploration collapse, recent research has explored incorporating dense supervision or off-policy guidance. While Process Reward Models (PRMs) have demonstrated efficacy in providing reliable step-wise verification for test-time scaling~\cite{zhang2025lessons}, \textbf{naively integrating PRM rewards into RLVR as dense rewards proves ineffective}~\cite{ye2025beyond}.A representative line of work that incorporate off-policy guidance is implementing Supervised Fine-Tuning (SFT) with RL at the \emph{trajectory level}, either by selectively fine-tuning hard instances or blending expert demonstrations with reinforcement objectives, as in ReLIFT, PrefixRFT, and BREAD~\cite{ma2025learning,huang2025blending,zhang2025bread}. Other methods impose \emph{structural or heuristic constraints} on reasoning, such as scaffolded optimization, step-level hints, or critique signals, to bias exploration toward higher-quality trajectories, including LUFFY, Scaf-GRPO, and Critique GRPO~\cite{yan2025learning,zhang2025scaf,zhang2025critique}. However, prior approaches rely on trajectory replacement or external supervision without performing explicit step-wise correction within model-generated rollouts. As a result, the substituted trajectories induce a distribution shift relative to the policy’s original rollouts, which not only leads to \textbf{insufficient exploration} of the valid reasoning space but also decreases sample efficiency by introducing misaligned rollouts.

To address the limitations of sparse outcome rewards and the resulting collapse in exploration diversity and sample efficiency, we propose \textbf{SCOPE} (\textbf{S}tep-wise \textbf{C}orrection for \textbf{O}n-\textbf{P}olicy \textbf{E}xploration), a novel framework designed to salvage partially correct trajectories and maximize exploration efficiency.Unlike methods that discard imperfect rollouts, SCOPE employs a \textbf{distribution-aware PRM selection strategy} to mitigate outlier instability and extract valid on-policy segments. SCOPE then applies fine-grained, step-wise off-policy rectification to surgically repair errors while preserving the correct context form policy model. By applying this targeted refinement, our method effectively transforms zero-reward rollouts into high-value training signals. Crucially, SCOPE actively sustains a diverse exploration space by generating varied, rectified reasoning paths that remain distributionally aligned with the model's original generation distribution, effectively counteracting the mode collapse observed in standard RLVR process.

Our contributions are summarized as follows:
\begin{itemize}[topsep=4pt, itemsep=2pt, parsep=0pt, partopsep=0pt]
        \item We introduce \textbf{SCOPE}, a framework that effectively integrates Process Reward Models into RLVR to \textbf{recycle valid reasoning steps} from failed rollouts. By implementing a distribution-aware selection strategy, we mitigate the risks of PRM instability on statistical outliers, enabling the robust extraction of informative on-policy segments that are typically discarded by outcome-based supervision.
    \item We demonstrate that SCOPE significantly improves \textbf{sample efficiency} through \textbf{precise, off-policy guided step-wise rectification}. By strictly targeting the first erroneous step while preserving the valid context, we transform zero-reward trajectories into high-quality training signals, resolving the credit assignment ambiguity inherent in sparse rewards.
    \item We conduct extensive experiments showing that SCOPE establishes new state-of-the-art performance, achieving 46.6\% average accuracy on in-distribution benchmarks and exhibiting robust generalization with 53.4\% accuracy on out-of-distribution tasks. Further analysis confirms that \textbf{SCOPE increases output diversity} by 13.5\% compared to GRPO, preventing mode collapse and sustaining a broad exploration space.
\end{itemize}

\begin{figure}[htbp]
    \centering
        \centering
        \includegraphics[width=\linewidth]{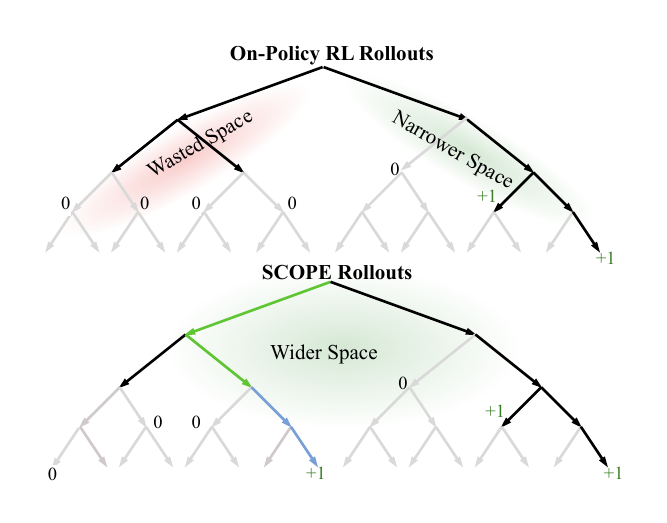}

        \label{fig:treeplot}
        \caption{\textbf{SCOPE Overview.}
        \textbf{Top:} On-policy RL rollouts terminate after the first error, causing partially correct trajectories to be discarded under sparse rewards.
        \textbf{Bottom:} SCOPE uses a PRM to identify the first incorrect step and applies step-wise off-policy rectification, while reusing correct on-policy steps to continue the rollout.
        \textbf{Colored Lines:} Black:Correct steps; Gray:Incorrect steps; Green:Recycled on-policy steps; Blue:Off-policy rectified steps.}

\end{figure}

\section{Related Work}

\subsection{Reinforcement Learning for LLM Reasoning}
The integration of Reinforcement Learning (RL) into Large Language Models (LLMs) has become a dominant paradigm for enhancing mathematical reasoning. Foundational PPO-based approaches~\cite{schulman2017proximal}, exemplified by DeepSeek-R1 and DeepSeekMath~\cite{guo2025deepseek,shao2024deepseekmath}, demonstrate substantial accuracy gains on complex benchmarks. To facilitate reproducibility, frameworks such as Open-R1, SimpleRL-Zoo, and DAPO have established scalable pipelines for RL training~\cite{hu2025open,zeng2025simplerl,yu2025dapo}.Concurrently, studies including AlphaMed and Logic-RL explore RL as a standalone optimization paradigm~\cite{liu2025alphamed,xie2025logicrl}, while DeepScaler provides empirical evidence that scaling RL training consistently outperforms supervised fine-tuning~\cite{luo2025deepscaler}. Despite these advances, the underlying mechanisms remain debated. Challenges such as reward hacking~\cite{skalse2022defining}, instability in zero-shot regimes~\cite{liu2025drgrpo}, and ambiguity regarding whether RL induces novel capabilities or merely amplifies pre-trained patterns persist~\cite{yue2025does,gandhi2025cognitive}. A critical limitation in sparse-reward formulations is the inability to distinguish partial correctness. Addressing this, recent works introduce process-level supervision: Light-R1 and VAPO employ curriculum strategies~\cite{wen2025lightr1,yue2025vapo}, TANGO utilizes generative verifiers~\cite{zha2025rltango}, and PRIME models implicit process rewards~\cite{cui2025prime}. Unlike these methods, our approach explicitly exploits intermediate errors to construct high-advantage trajectories.

\subsection{Process Reward Models}
Process Reward Models (PRMs) address the opacity of outcome supervision by evaluating intermediate reasoning steps. Foundational datasets such as PRM800K and Math-Shepherd demonstrate that step-level verification significantly outperforms outcome-based supervision~\cite{lightman2023let,pei2024shepherd,li2023making}, particularly when leveraging Chain-of-Thought structures~\cite{sprague2025to}. Subsequent research enhances expressiveness through next-token prediction and generative reasoning, as seen in Generative Verifiers and GenPRM~\cite{zhang2024generative,zhao2025genprm}, or by integrating explicit reasoning into verification, as in R-PRM and ThinkPRM~\cite{she2025rprmreasoningdrivenprocessreward,khalifa2025}. Parallel efforts focus on diagnostics and efficiency. Benchmarks such as ProcessBench, fine-grained error taxonomies, and robustness studies provide systematic insights~\cite{zheng2024processbench,pala2025errortypingsmarterrewards,zhang2025lessons}. Scaling research investigates inference-time compute trade-offs~\cite{wu2025inference}, annotation cost reduction via Free Process Rewards~\cite{yuan2025free}, and iterative preference learning~\cite{xiong2024iterative}. Furthermore, empirical studies indicate that smaller models rely heavily on process supervision for self-correction~\cite{zhang2024small}. In contrast to standard applications restricted to inference-time re-ranking, we utilize PRMs to actively identify erroneous steps for training signal construction.

\subsection{RLVR with Off-Policy Guidance}
While RLVR methods such as GRPO have demonstrated efficacy in reasoning alignment, they remain fundamentally constrained by inherent sample inefficiency and exploration instability. To mitigate these limitations, recent research has increasingly adopted off-policy guidance to regularize training dynamics, a direction substantiated by scalable infrastructures like DAPO and the theoretical analysis within LUFFY~\cite{yu2025dapo,yan2025learning}. One prevalent paradigm bridges Supervised Fine-Tuning (SFT) and RL through trajectory integration; methods such as ReLIFT, PrefixRFT, and BREAD selectively target hard instances or anchor exploration to expert demonstrations, effectively biasing the policy search toward high-quality regions~\cite{ma2025learning,huang2025blending,zhang2025bread}. Parallel efforts focus on injecting fine-grained structural priors: Scaf-GRPO and StepHint utilize scaffolded optimization and stepwise hints to shape reasoning paths~\cite{zhang2025scaf,zhang2025stephint}, while Critique-GRPO augments sparse scalar signals with informative natural language critiques~\cite{zhang2025critique}. Crucially, however, these approaches typically operate at the prompt or full-trajectory level. In contrast, our framework introduces \textbf{step-level rectification}, employing a PRM to precisely localize erroneous tokens and a stronger model to selectively rewrite them, thereby synthesizing high-granularity, high-advantage training data.

\section{Method}
\label{sec:method}
\begin{figure*}[t]
  \centering
  \includegraphics[width=\textwidth]{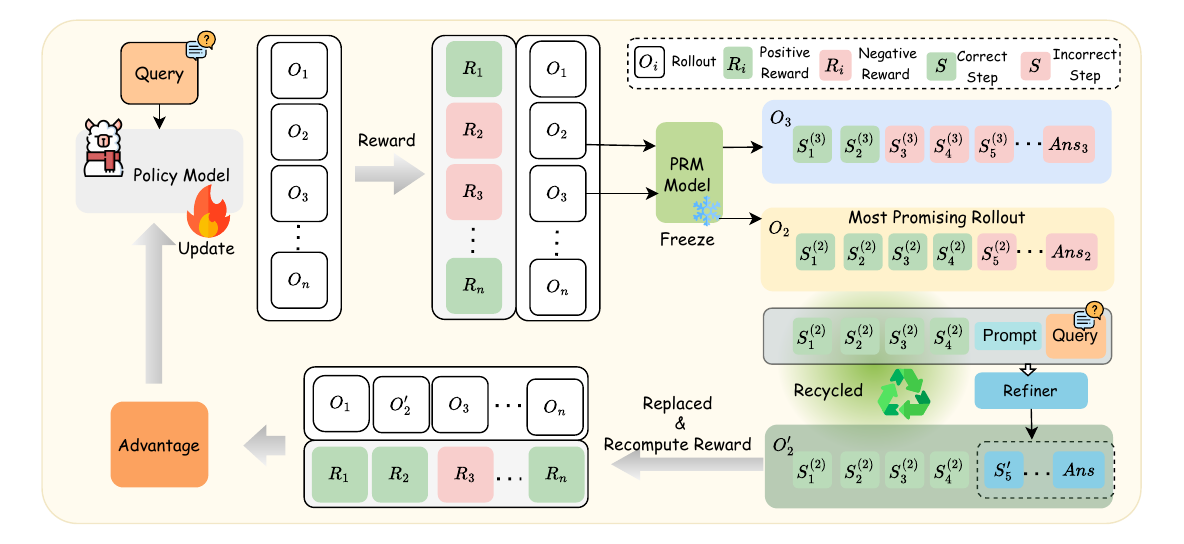}
  
 \caption{\textbf{Rollout recycling for sparse binary rewards.} For each query, the policy samples rollouts $\{O_i\}_{i=1}^n$ with rewards $\{R_i\}$. A PRM provides step scores $\{S_t\}$ to pick the most promising failed trajectory, yielding $O_2'$ by freezing its correct prefix and regenerating the erroneous suffix with a refiner. $O_2'$ is merged with on-policy rollouts and rewards are recomputed to preserve useful prefixes.We provide case studies in Appendix~\ref{Appendixexample} to illustrate how our method improves the model's suffix completions.}

  \label{fig:main}
\end{figure*}

We propose \textbf{SCOPE} (Step-wise Correction for On-Policy Exploration), a framework designed to mitigate credit leakage in binary-reward reasoning tasks. Standard RL often penalizes long, partially correct reasoning chains due to several downstream errors. SCOPE addresses this by identifying the precise boundary of valid reasoning using a Process Reward Model (PRM) and rectifying the subsequent error via an off-policy teacher, thereby constructing high-quality mixed training signals.

\subsection{PRM-Driven Error Localization and Selection}
\label{subsec:prm_selection}
Given a query $x$, we first sample $n$ rollouts $\mathcal{O}=\{O_1,\dots,O_n\}$ from the policy model $\pi_\theta$ and evaluate each trajectory with a scalar reward $R(O_i)$ (e.g., $R\in\{0,1\}$). We invoke PRM-driven localization only when there are at least two failing rollouts, i.e., $\left|\{\, i\in\{1,\dots,n\}\mid R(O_i)=0 \,\}\right|\ge 2$; otherwise we skip PRM selection(This is because in group-normalized GRPO with $\{0,1\}$ rewards, a single failure provides the only within-group contrast, and converting it to success collapses the reward variance and drives advantages toward zero.). To effectively salvage failing trajectories, we then employ a Process Reward Model (PRM) to pinpoint the first step-wise error among the zero-reward trajectories and select the most promising candidate for rectification.
\paragraph{Step-wise Verification.}
For a given prompt $x$, the policy $\pi_\theta$ generates a group of rollouts $\{y^{(i)}\}_{i=1}^G$. We decompose each rollout $y$ into a sequence of newline-delimited reasoning steps $(s_1, \dots, s_M)$. A PRM then scores each step, yielding probabilities $p_m \in [0, 1]$ representing the likelihood of correctness. We determine the \textit{verified-correct prefix} length $k$ as the index of the last step before the PRM confidence drops below a validity threshold $\tau$:
\begin{equation}
    k \;=\; \max \{0,m \mid p_1 \ge \tau, \dots, p_m \ge \tau \}.
\end{equation}
This boundary $k$ separates the valid on-policy reasoning (steps $s_{1:k}$) from the potential error (step $s_{k+1}$). We set the validity threshold to $\tau=0.5$ for correctness in our experiments..

\paragraph{Distribution-Aware Selection Strategy.}
To address the computational inefficiency and stability risks associated with rectifying statistical outliers, we propose a distribution-aware scoring mechanism that balances reasoning progress against statistical representativeness. Within the subset of rollouts receiving a terminal reward of $0$, we select the candidate $O_i$ that maximizes a weighted score $\mathcal{S}(O_i)$.

First, for each rollout $O_i$ in the batch, we determine the ratio of verified reasoning to total length. Let $k_i$ be the number of verified correct steps and $M_i$ be the total step count. Similarly, utilizing the tokenizer, let $c_i$ be the token count of the verified prefix and $L_i$ be the total token count. We calculate the progress ratios:
\begin{equation}
    r_{\text{step}}^{(i)} = \frac{k_i}{M_i}, \quad r_{\text{token}}^{(i)} = \frac{c_i}{L_i}.
\end{equation}

Second, to penalize extreme outliers (e.g., super-long trajectories), we compute the batch-wise mean and variance for steps ($\mu_M, \sigma_M^2$) and tokens ($\mu_L, \sigma_L^2$). We derive Gaussian weights $\alpha_i$ and $\beta_i$ based on the rollout's deviation from the batch distribution:
\begin{equation}
    \alpha_i = \exp\left(-\frac{(M_i - \mu_M)^2}{2\sigma_M^2}\right), \quad \beta_i = \exp\left(-\frac{(L_i - \mu_L)^2}{2\sigma_L^2}\right).
\end{equation}
These weights equal $1$ when the length aligns with the batch mean and decay probabilistically as the length deviates.

Finally, the selection score $\mathcal{S}(O_i)$ is computed as the product of these distribution weights and the sum of the progress ratios:
\begin{equation}
    \mathcal{S}(O_i) = \alpha_i \cdot \beta_i \cdot \left( r_{\text{step}}^{(i)} + r_{\text{token}}^{(i)} \right).
\end{equation}
We select the rollout $O^*$ with the highest score $\mathcal{S}(O^*)$. This strategy ensures we prioritize "near-miss" trajectories that are distributionally representative, effectively avoiding computational hazards associated with extreme length outliers.

\subsection{SCOPE: Step-wise Correction for On-Policy Exploration}
\label{subsec:scope_training}
Once a target prefix $y_{\le L}$ (corresponding to steps $s_{1:k}$) is identified, SCOPE constructs a rectified trajectory and applies a hybrid optimization objective to leverage both on-policy and off-policy data.

\paragraph{Step-Wise Off-Policy Guided Rectification.}
We prompt an external teacher model $\mu$ to generate a correct continuation conditioned on the preserved student prefix. The teacher produces a sequence of rectified steps $S'_{>k}$ using deterministic decoding to ensure stability. We then synthesize a mixed trajectory $O'_i$ by concatenating the student's valid recycled steps with the teacher's corrected suffix:
\begin{align}
    O'_i \;=\; \mathrm{concat}(S_{1:k}^{(i)}, S'_{>k}).
\end{align}
This process creates a high-advantage training sample $O'_i$ (labeled as $O'_2$ in Figure~\ref{fig:main}) where the prefix $S_{1:k}^{(i)}$ strictly follows the student's current policy distribution ($\pi_\theta$), while the suffix $S'_{>k}$ acts as an off-policy guide. We define a binary mask $m_t$ to distinguish between tokens belonging to the recycled on-policy steps ($m_t=0$) and the refined off-policy steps ($m_t=1$).

\paragraph{Hybrid Optimization Objective.}
Extending from the standard GRPO objective, we formulate a hybrid objective to leverage the dual nature of the rectified rollout $O'_i$. Let $y = (y_1, \dots, y_T)$ denote the token sequence of $O'_i$, where the first $k$ steps are recycled on-policy tokens ($S_{1:k}^{(i)}$) and the subsequent steps are off-policy teacher corrections ($S'_{>k}$).

To rigorously combine these signals, we define the objective $\mathcal{J}_{\text{SCOPE}}(\theta)$ as a maximization problem. For the on-policy prefix, we employ the standard PPO-style clipped surrogate objective to stabilize updates. For the off-policy suffix, we utilize a weighted likelihood objective that encourages the policy to clone the teacher's reasoning without being hindered by the advantage estimation variance inherent in RL. The formulation is as follows:
\begin{equation}
    \label{eq:scope_obj}
    \begin{split}
        \mathcal{J}_{\text{SCOPE}}(\theta) &= \frac{1}{Z} \Bigg( \underbrace{\sum_{t=1}^{k} \text{CLIP}(r_{t}(\theta), \hat{A}_i, \epsilon)}_{\text{On-Policy Prefix (RL)}} \\
        &\quad + \underbrace{\rho \sum_{t=k+1}^{T} w_\gamma(p_t) \log \pi_\theta(y_t \mid y_{<t}) \cdot \hat{A}_i}_{\text{Off-Policy Suffix}} \Bigg),
    \end{split}
\end{equation}
\noindent where $Z = T$ serves as the length normalization factor, and $\rho$ controls the weight of the off-policy guidance.
The term $r_{t}(\theta) = \frac{\pi_\theta(y_t \mid y_{<t})}{\pi_{\theta_{\text{old}}}(y_t \mid y_{<t})}$ represents the probability ratio for importance sampling on the prefix.
The clipping function is defined as $\text{CLIP}(r, A, \epsilon) = \min(rA, \text{clip}(r, 1-\epsilon, 1+\epsilon)A)$, which prevents destructive updates. The shaping weight is defined as $w_\gamma(p_t) = \frac{p_t}{p_t + \gamma}$, where $p_t = \pi_\theta(y_t \mid y_{<t})$ and $\gamma$ is a hyperparameter (following \cite{yan2025learning}, we set $\gamma=1$).

\section{Experiments}
\subsection{Experimental Setup}

\paragraph{{Training Dataset.}}
We follow the dataset setup used in SimpleRL~\cite{zeng2025simplerl} and Dr.GRPO~\cite{liu2025oat}. Specifically, we construct the training set from the MATH~\cite{hendrycks2021math500} dataset by selecting problems with difficulty Levels 3–5, resulting in 8,523 training examples.Training details are provided in Appendix~\ref{training-details}.
\paragraph{{Models.}}
We evaluate our method on three representative backbones spanning both general purpose and math-specialized pretraining. Concretely, we run independent training and evaluation on Qwen2.5-1.5B~\cite{qwen2025qwen25technicalreport}, Qwen2.5-Math-7B~\cite{yang2024qwen2}, and Llama-3.1-8B~\cite{grattafiori2024llama}, allowing us to assess consistency across model families and scales. We use Qwen2.5-Math-PRM-7B~\cite{zhang2025lessons} as the \textbf{process reward model (PRM)} for step-level evaluation.(Details of the PRM model are provided in Appendix~\ref{prm-details}.)We use Deepseek-R1~\cite{guo2025deepseek} as the \textbf{Refiner} to provide step-wise off-policy guidance for recyclable rollouts and producing rectified trajectories for optimization.
\paragraph{Baseline Methods.}
We evaluate our method against three categories of baselines. \textbf{Previous RLVR Methods} include OpenReasoner-Zero~\cite{hu2025open}, PRIME-Zero~\cite{cui2025prime}, and Oat-Zero~\cite{liu2025oat}, alongside the Qwen-Base and Qwen-Instruct models; these represent standard on-policy reinforcement learning approaches. \textbf{Off-policy Learning Methods} consist of SFT and SFT+RL, representing strategies that directly incorporate supervised data into the learning objective. \textbf{Off-policy Guided RL Methods} include LUFFY~\cite{yan2025learning}, ReLIFT~\cite{ma2025learning}, and ScafGRPO~\cite{zhang2025scaf}, which utilize off-policy guidance such as anchors, hints, or scaffolds to stabilize training. 
For Qwen basemodels and Zero-suffix RLVR Methods, we report rusults on their publicly available checkpoints; for other RL Methods and Off-policy Guided RL Methods, we conduct experiment following their published optimal parameter settings on our dataset for fair comparison.

\paragraph{Evaluation Details.}
We assess in-distribution (ID) mathematical reasoning capabilities using the AIME24\cite{li2024numinamath}, AIME25\cite{li2024numinamath}, AMC\cite{li2024numinamath}, MATH-500\cite{hendrycks2021math500}, Minerva\cite{lewkowycz2022minerva}, and Olympiad benchmarks\cite{he2024olympiadbench} . For out-of-distribution (OOD) generalization, we utilize ARC-c\cite{clark2018arc}, GPQA-Diamond\cite{rein2024gpqa}, and MMLU-Pro\cite{wang2024mmlu}. Regarding limited number of samples, we report \texttt{Avg@32} over 32 independent runs for AIME24, AIME25, and AMC to ensure statistical stability, while \texttt{pass@1} is reported for all other benchmarks. For implementation, we evaluate \textbf{Qwen baselines} and \textbf{Previous RLVR Methods} using their publicly available weights. For \textbf{Off-policy Learning Methods} and \textbf{Off-policy Guided RL Methods}, we conduct experiments on our own training data using the hyperparameter settings specified in their respective original publications.

\definecolor{IDAvgColor}{HTML}{FFF2CC}   
\definecolor{OODAvgColor}{HTML}{E8F1FF}  

\definecolor{IDheadColor}{HTML}{D9FFDD}
\definecolor{OODheadColor}{HTML}{D9DFFF}
\definecolor{methodcolor}{gray}{0.93}

\begin{table*}[htbp]
\centering
\caption{Performance comparison (\%) on the Qwen2.5-Math-7B backbone. We benchmark SCOPE against three categories of baselines: previous RLVR approaches, off-policy learning strategies, and off-policy guided methods. The best performance for each task is highlighted in bold.}
\label{tab:id_ood_results}

\setlength{\tabcolsep}{1pt} 

\begin{tabular}{
l
cccccc >{\columncolor{IDAvgColor}}c
|
ccc >{\columncolor{OODAvgColor}}c
}
\toprule
\multirow{2}{*}{\textbf{Model}} 
& \multicolumn{7}{c|}{\cellcolor{IDheadColor}\textbf{In-Distribution Performance}} 
& \multicolumn{4}{c}{\cellcolor{OODheadColor}\textbf{Out-of-Distribution Performance}} \\
\cmidrule(lr){2-8} \cmidrule(lr){9-12}

& \textbf{AIME24} 
& \textbf{AIME25} 
& \textbf{AMC} 
& \textbf{MATH-500} 
& \textbf{Minerva} 
& \textbf{Olympiad} 
& \textbf{Avg.}
& \textbf{ARC-c} 
& \textbf{GPQA*} 
& \textbf{MMLU-Pro} 
& \textbf{Avg.} \\
\midrule

\textbf{Qwen-Base}
& 11.5 & 4.9 & 31.3 & 43.6 & 7.4 & 15.6 & 19.1
& 18.2 & 11.1 & 16.9 & 15.4 \\

\textbf{Qwen-Instruct}
& 12.5 & 10.2 & 48.5 & 80.4 & 32.7 & 41.0 & 37.6
& 70.3 & 24.7 & 34.1 & 43.0 \\
\midrule

\multicolumn{12}{l}{\cellcolor{methodcolor}\textbf{Previous RLVR Methods}} \\
\midrule
On-Policy RL
& 18.4 & 6.8 & 54.9 & 76.6 & 36.0 & 42.6 & 39.2
& 62.0 & 21.7 & 43.53 & 42.3 \\

OpenReasoner-Zero
& 16.5 & 15.0 & 52.1 & 82.4 & 33.1 & 47.1 & 41.0
& 66.2 & 29.8 & \textbf{58.7} & 51.6 \\

PRIME-Zero
& 17.0 & 12.8 & 54.0 & 81.4 & 39.0 & 40.3 & 40.8
& 73.3 & 18.2 & 32.7 & 41.4 \\

Oat-Zero
& 33.4 & 11.9 & 61.2 & 78.0 & 34.6 & 43.4 & 43.8
& 70.1 & 23.7 & 41.7 & 45.2 \\

\midrule

\multicolumn{12}{l}{\cellcolor{methodcolor}\textbf{Off-policy Learning Methods}} \\
\midrule
SFT
& 19.3 & 18.6 & 49.6 & 77.8 & 36.2 & 41.7 & 40.5
& 71.2 & 23.7 & 39.8 & 44.9 \\


SFT+RL
& 20.9 & 21.6 & 51.2 & 78.6 & 36.3 & 43.4 & 42.0
& 68.5 & 21.9 & 35.8 & 42.1 \\
\midrule

\multicolumn{12}{l}{\cellcolor{methodcolor}\textbf{Off-policy Guided RL Methods}} \\
\midrule
LUFFY
& 24.4 & \textbf{18.9} & 57.8 & 80.4 & 37.1 & 41.7 & 43.4
& 73.6 & 32.3 & 44.7 & 50.2 \\

ReLIFT
& 28.3 & 16.7 & 58.7 & 81.2 & 38.2 & 45.6 & 44.8
& 72.3 & 33.6 & 48.6 & 51.5 \\


Scaf-GRPO
& 30.2 & 11.8 & 58.1 & 80.6 & 34.6 & 43.0 & 43.0
& 69.6 & 31.2 & 44.3 & 48.4 \\
\midrule

\textbf{SCOPE(Ours)}
& \textbf{30.7} & 17.6 & \textbf{62.4} & \textbf{82.4} & \textbf{39.3} & \textbf{47.0} & \textbf{46.6}
& \textbf{73.4} & \textbf{34.6} & 52.1 & \textbf{53.4} \\

\bottomrule
\end{tabular}
\end{table*}

\subsection{Main Results}
\label{subsec:main_results}

Table~\ref{tab:id_ood_results} presents a comprehensive quantitative evaluation of our proposed method against three distinct categories of baselines: previous RLVR approaches, standard off-policy learning methods, and recent off-policy guided RL frameworks. 
\paragraph{In-Distribution Performance Analysis.}
The left part of Table~\ref{tab:id_ood_results} summarizes in-distribution results across all baselines. While standard RLVR methods improve over the Qwen-Instruct baseline, their performance varies substantially across benchmarks. In contrast, SCOPE exhibits consistently balanced performance across difficulty levels. It achieves the best results on \textbf{AMC} (62.4\%) and \textbf{Minerva} (39.3\%), ties for the top performance on MATH-500 (82.4\%), and remains competitive on Olympiad benchmarks (47.0\%). Overall, SCOPE attains the highest average in-distribution accuracy (46.6\%), exceeding the strongest competing method by +1.8\%. Compared to off-policy guided baselines that excel on individual tasks but generalize inconsistently, these results indicate that fine-grained, step-wise rectification yields a more stable and generalizable reasoning policy than trajectory-level replacement or coarse guidance.

\paragraph{{Out-of-Distribution Generalization.}}

The right part of Table~\ref{tab:id_ood_results} summarizes out-of-distribution results. While most baselines exhibit performance degradation under distribution shift, our method maintains the strongest overall generalization, achieving the highest average OOD accuracy (53.4\%). SCOPE attains the best performance on \textbf{ARC-c} (73.4\%) and \textbf{GPQA*} (34.6\%). Compared with the strongest off-policy guided baselines, our approach improves the OOD average by +3.2\% over LUFFY~\cite{yan2025learning} and +1.9\% over ReLIFT~\cite{ma2025learning}. These results indicate that step-level rectification does not overfit in-distribution benchmarks and transfers reliably to out-of-distribution reasoning tasks.
\begin{figure}[t]
  \centering
  \includegraphics[width=\columnwidth]{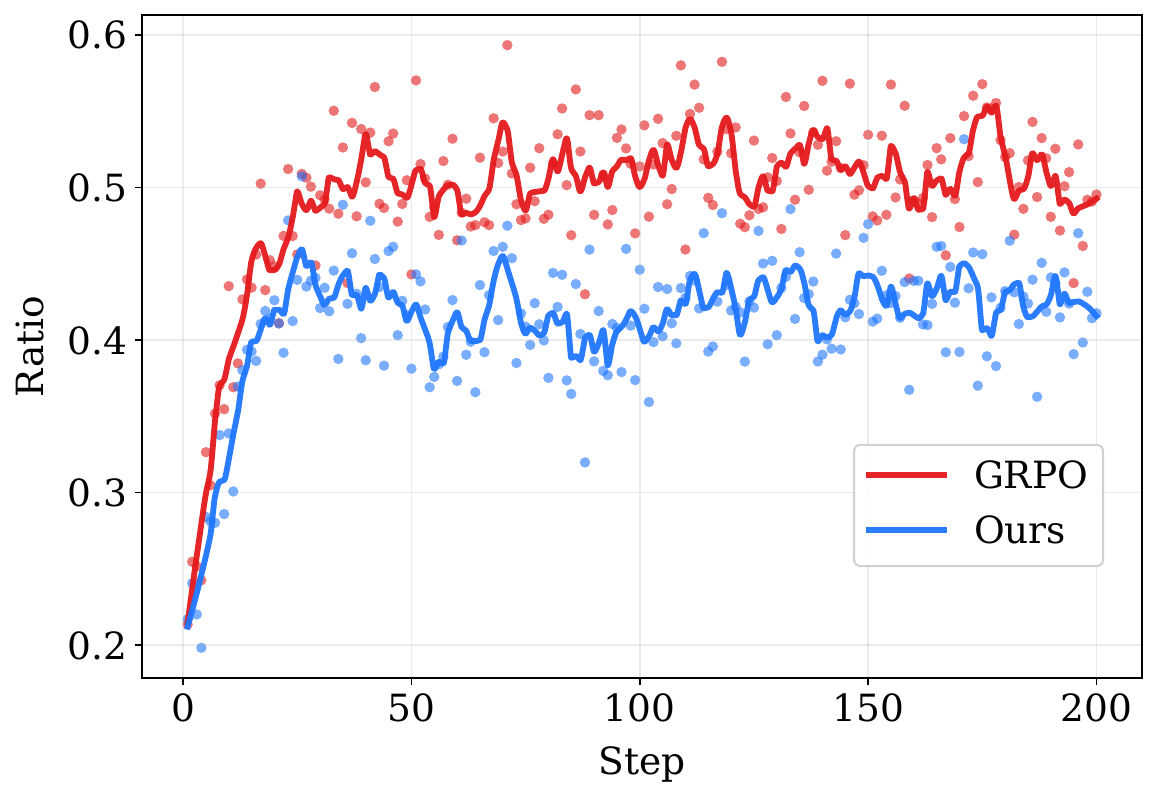}
  \caption{Discarded valid reasoning trajectories among zero-reward samples. Compared to GRPO, our method significantly reduces the fraction of potentially valid trajectories that are discarded.}
  \label{fig:discarded-traj}
\end{figure}
\paragraph{Generation Diversity and Quality.}
As shown in Fig.~\ref{fig:discarded-traj}, our method recycles usable reasoning trajectory from zero-reward trajectories, reducing the discard rate of potentially valid reasoning prefixes relative to GRPO and improving effective exploration(Additional experimental results and analyses are provided in Appendix~\ref{appendixreducenermiss}).To verify that SCOPE increases exploration rather than collapsing to a single mode, we measure reasoning-path diversity in Figure~\ref{fig:radar}. We report Distinct-$n$~\cite{chen2025diversity} and intra-set similarity metrics Self-BLEU~\cite{papineni2002bleu} / Self-ROUGE~\cite{lin2004rouge} (as $1-\text{Score}$, higher is better). SCOPE consistently improves all diversity axes over GRPO, with notably higher Distinct-1/2/4 and reduced redundancy in $1-\text{SelfBLEU}$ and $1-\text{SelfROUGE}$. Importantly, this broader exploration also improves quality (higher pass@1 and pass@10), indicating that step-wise rectification yields more diverse \emph{and} more successful reasoning trajectories(We report detailed diversity and quality results in Appendix~\ref{appendixdiversity}.).
\subsection{Training Dynamics}
\label{subsec:training_dynamics}

\begin{figure*}[t]
    \centering

    \begin{minipage}{\columnwidth}
        \centering

        \includegraphics[width=0.86\linewidth]{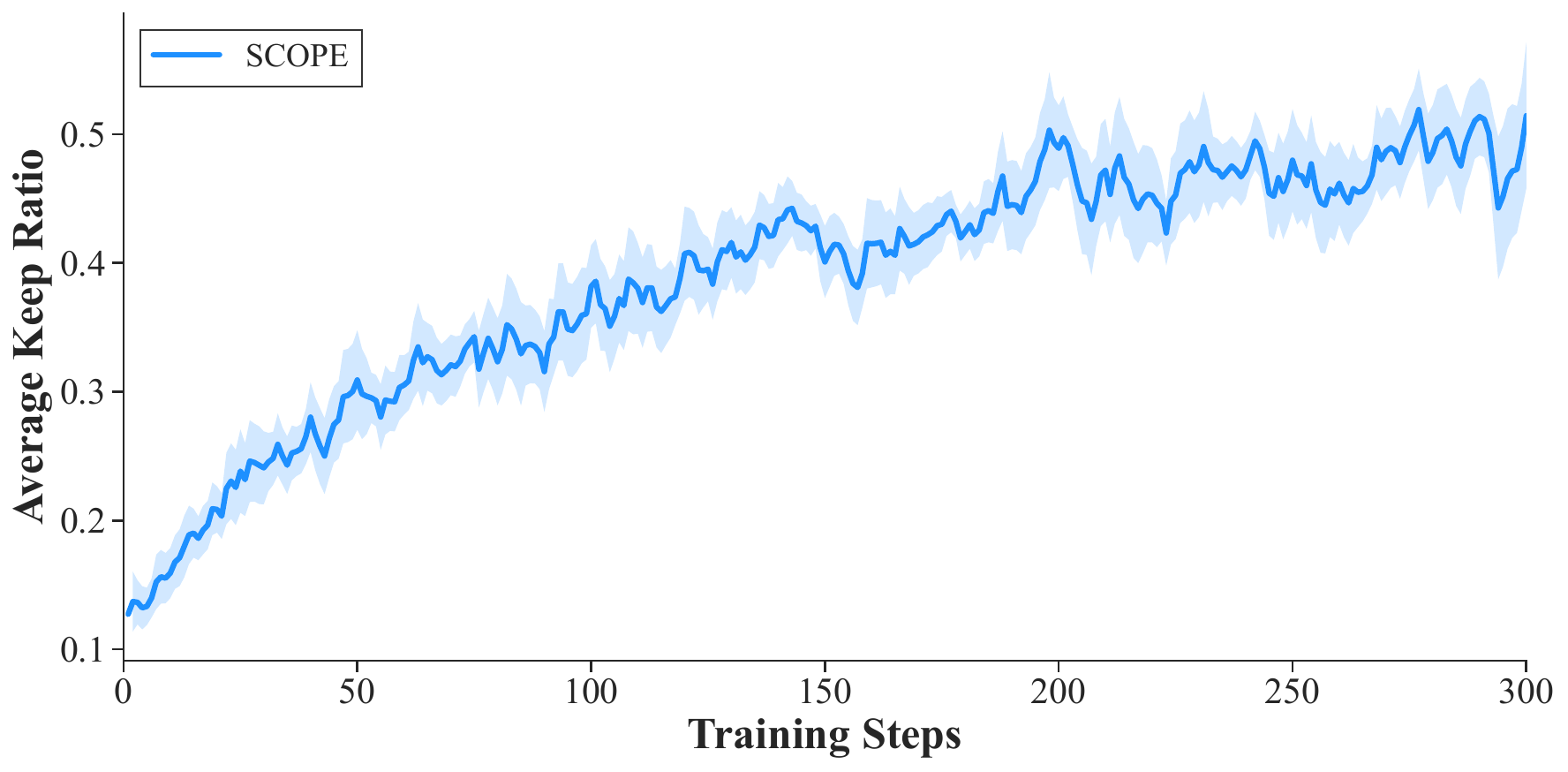}
        \subcaption{Avg. Keep Ratio}
        \label{fig:keep_ratio}
    \end{minipage}%
    \hfill 
    \begin{minipage}{\columnwidth}
        \centering
        \includegraphics[width=0.86\linewidth]{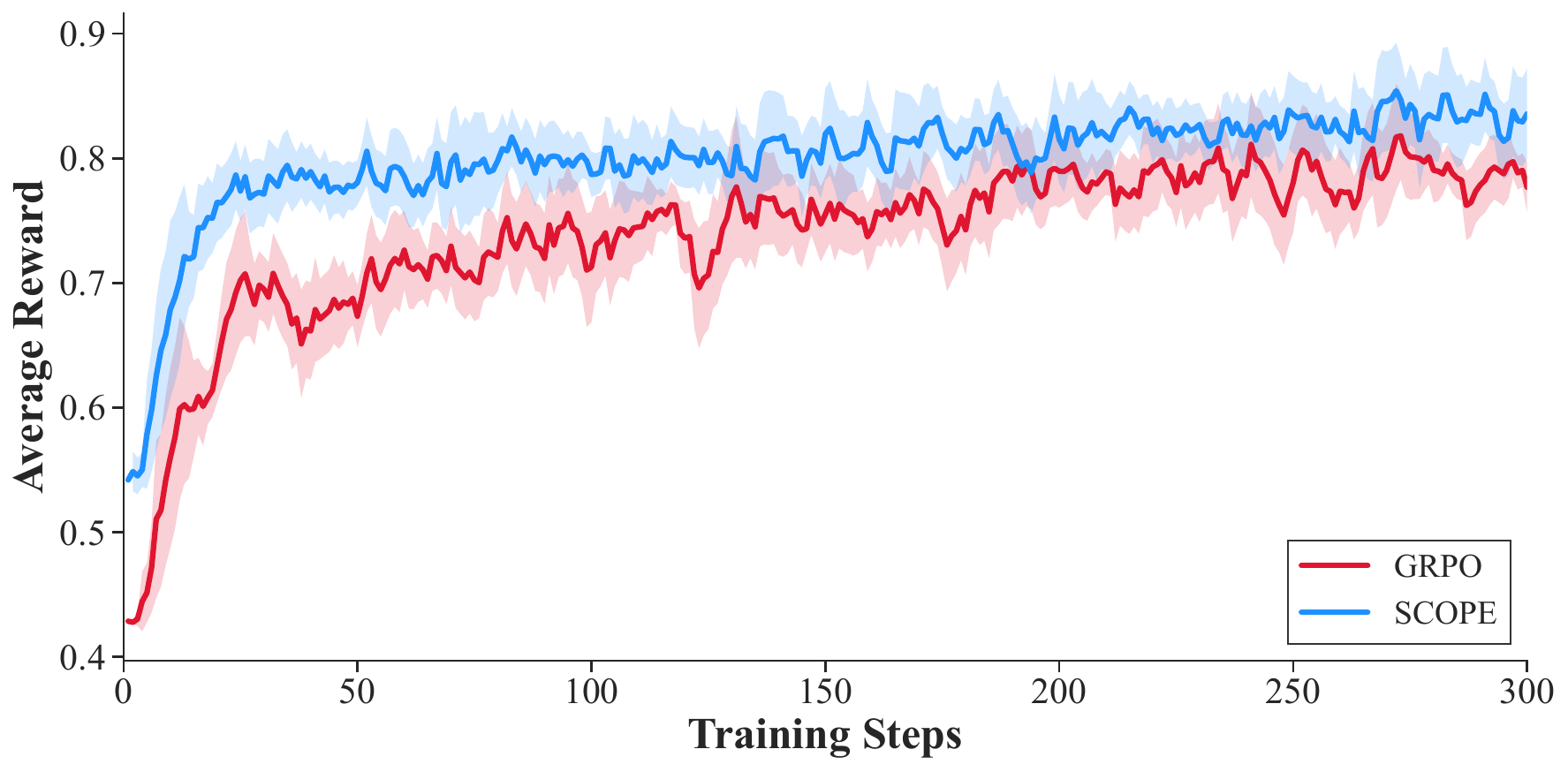}
        \subcaption{Reward Curves}
        \label{fig:reward}
    \end{minipage}

    \vspace{1em}

    \begin{minipage}[b]{\columnwidth}
        \centering
        \includegraphics[width=0.86\linewidth]{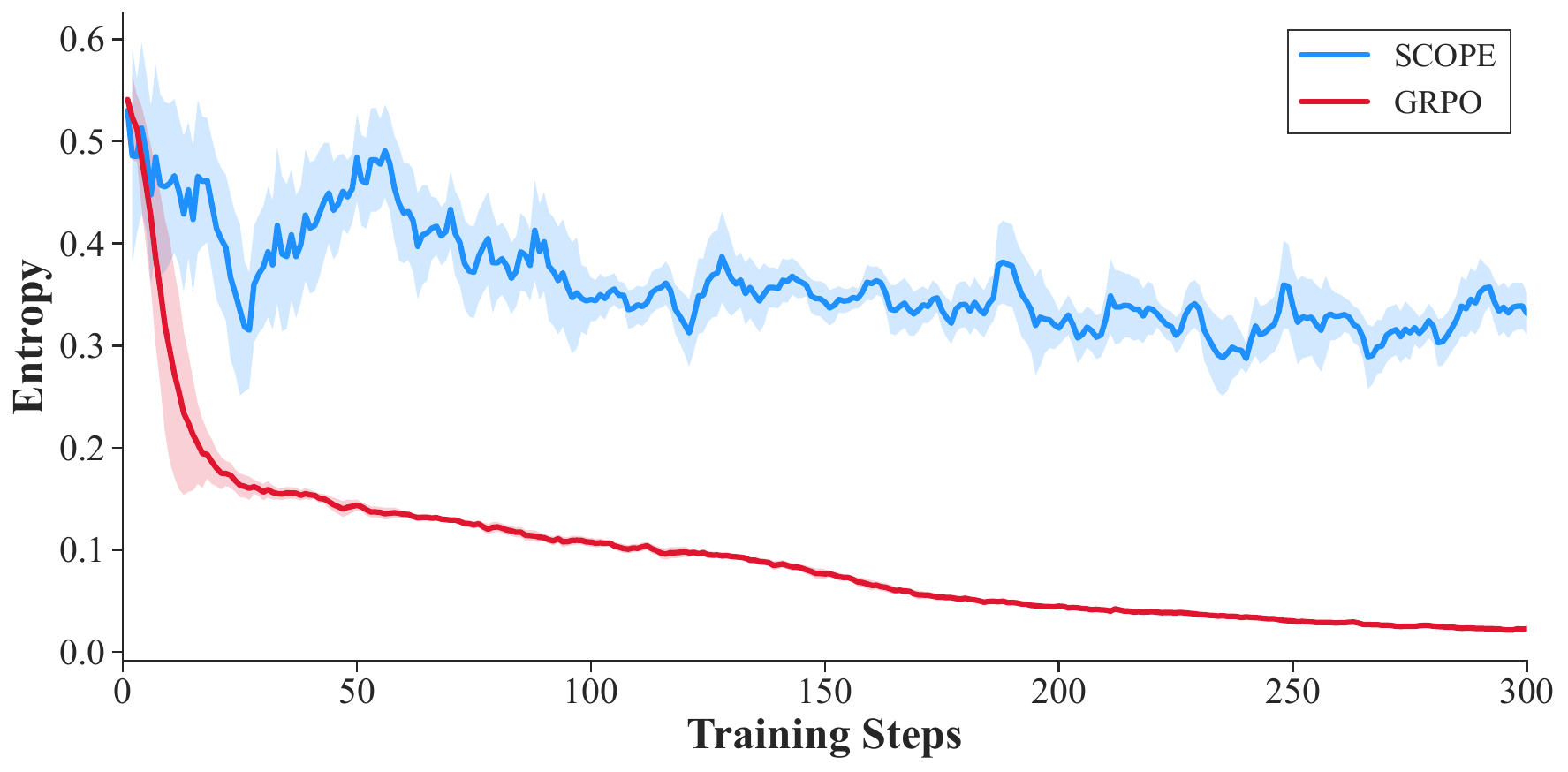}
        \subcaption{Policy Entropy}
        \label{fig:entropy}
    \end{minipage}%
    \hfill %
    \begin{minipage}[b]{\columnwidth}
        \centering
        \includegraphics[width=0.96\linewidth]{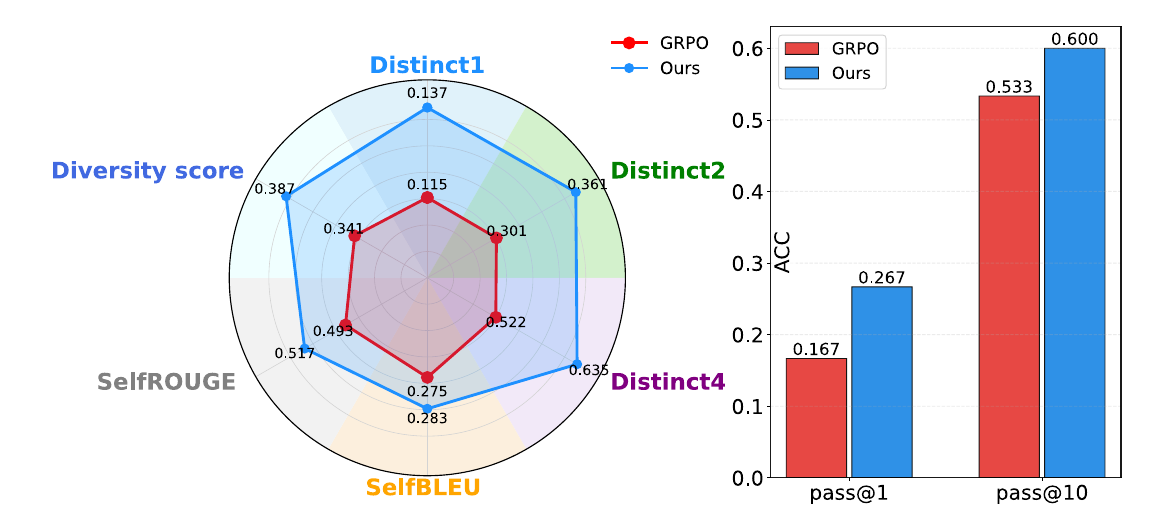} 
        \subcaption{Diversity and Quality Analysis}
        \label{fig:radar}
    \end{minipage}

    \caption{\textbf{Training Dynamics Analysis.} We present the training curves for (a) the average step keep ratio during rectification, (b) the reward comparison between our method and GRPO, (c) the policy entropy, and (d) the diversity and quality results from 10-sample evaluations on the AIME24 dataset at temperature $T{=}1$.}
    \label{fig:training_dynamics}
\end{figure*}

To empirically validate the stability and optimization efficiency of our proposed method, we visualize the evolution of key training metrics in Figure~\ref{fig:training_dynamics}. These metrics provide insights into the model's behavior regarding step-level rectification, reward maximization, exploration maintenance, and reasoning complexity(We additionally report further training dynamics, including training- and validation-set accuracy as well as response length, in Appendix~\ref{appendixtraingdyanmic}.).




\paragraph{Effectiveness of step-level Rectification.}
Figure~\ref{fig:keep_ratio} illustrates the average keep ratio of steps retained by the Process Reward Model (PRM) during rectification. The ratio increases steadily over training, which indicates our training policy progressively produces fewer tokens requiring correction. This result demonstrates that step-level feedback is effectively incorporated into the learning process, supporting the intended fine-grained optimization.

\paragraph{Reward Optimization.}
Figure~\ref{fig:reward} compares the reward trajectories of our method with the standard GRPO baseline. Our approach achieves consistently higher rewards throughout training, reflecting more stable and efficient learning, which proves that leveraging rectified, high-advantage trajectories enhances the signal quality and improves sample efficiency.

\paragraph{Exploration Maintenance.}
Figure~\ref{fig:entropy} shows the evolution of policy entropy during training. Our method maintains a relatively high entropy level with periodic fluctuations, indicating continued exploration throughout learning. Sustained entropy helps prevent premature convergence, allowing the policy to explore diverse reasoning trajectories in complex mathematical domains.


\subsection{Extension to More Models}
\definecolor{methodcolor}{gray}{0.93}

\begin{table*}[ht]
\centering
\setlength{\tabcolsep}{6pt}  
\renewcommand{\arraystretch}{1.1}
\begin{tabular}{lccccccc}
\toprule
\textbf{Model} & \textbf{AIME24} & \textbf{AIME25} & \textbf{AMC} & \textbf{MATH-500} & \textbf{Minerva} & \textbf{Olympiad} & \textbf{Avg.} \\
\midrule

\multicolumn{8}{c}{\cellcolor{methodcolor}\textbf{Qwen2.5-1.5B }} \\
\midrule
Qwen2.5-1.5B-Base
  & 0.0 & 1.0 & 1.5 & 4.0 & 1.8 & 1.8 & 1.7 \\
  Qwen2.5-1.5B-Instruct
  & 2.3 & 0.9 & 21.2 & 49.4 & 15.8 & 16.4 & 17.7 \\
On-Policy RL
  & \textbf{4.7} & 0.0 & 17.2 & 32.6 & 6.3 & 13.3 & 12.3 \\
SCOPE
  & 4.1 & \textbf{1.7} & \textbf{24.7} & \textbf{57.0} & \textbf{19.9} & \textbf{21.5} & \textbf{21.5} \\
  
\midrule

\multicolumn{8}{c}{\cellcolor{methodcolor}\textbf{Llama3.1-8B }} \\
\midrule
LLaMA-3.1-8B-Base
  & 0.0 & 0.0 & 4.7 & 11.0 & 4.4 & 1.9 & 3.7 \\
On-Policy RL
  & 0.0 & 0.1 & 1.2 & 17.4 & 1.8 & 0.6 & 3.5 \\
SCOPE
  & 0.0 & \textbf{2.0} & \textbf{7.6} & \textbf{29.8} & \textbf{15.8} & \textbf{7.6} & \textbf{10.5} \\
\bottomrule
\end{tabular}
\caption{Scalability analysis on Qwen2.5-1.5B and Llama-3.1-8B. SCOPE consistently outperforms standard on-policy RL and stabilizes training across different model architectures where standard RL often fails.}
\label{tab:other_model_table}
\end{table*}
To evaluate scalability across model sizes, we further conduct experiments on Qwen2.5-1.5B and Llama3.1-8B. Table~\ref{tab:other_model_table} reports in-distribution performance on mathematical reasoning benchmarks.

\paragraph{Qwen2.5-1.5B.}
Standard on-policy RL shows unstable on Qwen2.5-1.5B and underperforms the Instruct baseline (-5.4\%). SCOPE stabilizes optimization and improves average accuracy to 21.5\% (+3.8\%), with strong gains on MATH-500 (+7.6\%) and AMC (+3.5\%), indicating recovered learning signal for small models.

\paragraph{LLaMA3.1-8B.}
On LLaMA3.1-8B, on-policy RL yields no improvement over the base model (3.7\% vs.\ 3.5\%). SCOPE boosts average accuracy to 10.5\% (+6.8\%), with notable gains on MATH-500 (+12.4\%) and Minerva (+14.0\%), suggesting step-wise off-policy rectification generalizes to mid-scale models.
\begin{table*}[ht]
\centering
\setlength{\tabcolsep}{6pt}
\renewcommand{\arraystretch}{1.1}
\begin{tabular}{lccccccc}
\toprule
\textbf{Model} 
& \textbf{AIME24} 
& \textbf{AIME25} 
& \textbf{AMC} 
& \textbf{MATH-500} 
& \textbf{Minerva} 
& \textbf{Olympiad} 
& \textbf{Avg.} \\
\midrule
On-policy RL
& 18.4 & 6.8 & 54.9 & 76.6 & 36.0 & 42.6 & 39.2 \\
Replace Entire Rollout
& 24.4 & \textbf{18.9} & 57.8 & 80.4 & 37.1 & 41.7 & 43.4 \\
RL w/PRM Reward
& 12.9 & 7.2 & 52.5 & 78.6 & 34.5 & 40.4 & 37.7 \\
Qwen2.5-32B-Instruct as Refiner
& 23.8 & 16.6 & 60.6 & 81.2 & \textbf{42.3} & 43.4 & 44.6 \\
\textbf{SCOPE(Ours)}
& \textbf{30.7} & 17.6 & \textbf{62.4} & \textbf{82.4} & 39.3 & \textbf{47.0} & \textbf{46.6}\\
\bottomrule
\end{tabular}
\caption{Ablation study on refinement strategies and reward integration. Results demonstrate that SCOPE's step-wise rectification significantly outperforms whole-trajectory replacement and naive PRM reward densification.}
\label{tab:ablation_rollout_refiner}
\end{table*}

\subsection{Ablation on Refinement Strategies}
\label{subsec:ablation_rollout}
Table~\ref{tab:ablation_rollout_refiner} reports our ablation study comparing different strategies for leveraging reward signals and off-policy guidance.




\paragraph{Naive PRM Reward Integration Fails.}
The comparison between \textbf{On-policy RL} and \textbf{RL w/ PRM Reward} shows that directly injecting step-wise PRM scores into the reward function degrades performance, reducing the average accuracy from 39.2\% to 37.7\%. This result indicates that naively densifying sparse rewards introduces misaligned optimization signals, underscoring the need for structured use of PRM feedback rather than direct reward shaping.

\paragraph{Step-wise Rectification Outperforms Whole-Trajectory Replacement.}
The comparison between \textbf{Replace Entire Rollout} and \textbf{SCOPE (Ours)} reveals a clear advantage of step-wise rectification. While whole-trajectory replacement improves average accuracy to 43.4\%, SCOPE further raises performance to 46.6\%. SCOPE preserves the correct prefix produced by the student model and applies off-policy guidance only at the first erroneous step, keeping the learning signal locally aligned with the model’s reasoning state and enabling more effective policy updates.

\paragraph{Robustness to Refiner Model Quality.}
Using \textbf{Qwen2.5-32B-Instruct as Refiner}~\cite{team2024qwen2} achieves 44.6\% accuracy, outperforming the \textbf{Replace Entire Rollout} baseline despite the weaker teacher. This result validates our core contribution: \textbf{expanding the exploration space while strictly maintaining distributional alignment}. By retaining the valid on-policy prefix, SCOPE ensures that corrections remain grounded in the model's current distribution, transforming off-policy signals into effective exploration targets. This proves that minimizing distribution shift is more critical than the absolute capability of the teacher.
\section{Conclusion}

In this work, we present \textbf{SCOPE}, a reinforcement learning framework that addresses the fundamental challenges of sample inefficiency and exploration instability in reasoning tasks.By integrating Process Reward Models with a novel step-wise rectification structure, SCOPE transforms previously discarded, near-correct trajectories into high-value training signals while preserving the distributional consistency of the policy’s original reasoning context. SCOPE's contribution that maximizing data efficiency through fine-grained error correction and stabilizing alignment by maintaining on-policy prefixes enables significantly faster convergence and more robust optimization than traditional off-policy guided RLVR methods. Our extensive evaluation demonstrates that SCOPE establishes new state-of-the-art performance across in-distribution and out-of-distribution benchmarks while effectively increasing generation diversity, thereby preventing mode collapse and unlocking reasoning capabilities across models where standard reinforcement learning often falters.
\section*{Impact Statement}
This paper presents work whose goal is to advance the field of Machine
Learning. There are many potential societal consequences of our work, none
which we feel must be specifically highlighted here.
\bibliography{example_paper}
\bibliographystyle{icml2026}

\newpage
\onecolumn
\appendix

\section{Additional Results}
\subsection{Diversity and Quality}
\label{appendixdiversity}
\begin{table}[ht]
\centering
\small
\setlength{\tabcolsep}{4pt}
\begin{tabular}{lccccc cc}
\toprule
\multirow{2}{*}{Method} &
\multicolumn{5}{c}{Diversity $\uparrow$ (p50 [p10, p90])} &
\multicolumn{2}{c}{Quality $\uparrow$} \\
\cmidrule(lr){2-6}\cmidrule(lr){7-8}
& distinct-1 & distinct-4 & 1-self-bleu & 1-self-rouge & div-score
& pass@1 & pass@10 \\
\midrule
\multicolumn{8}{l}{\textbf{AIME24} ($K{=}10$ samples per prompt)} \\
\midrule
GRPO (baseline) &
0.11 [0.07, 0.16] &
0.52 [0.34, 0.68] &
0.27 [0.17, \textbf{0.41}] &
0.49 [\textbf{0.37}, \textbf{0.63}] &
0.34 [0.27, 0.41] &
16.67 & 53.33 \\
Ours &
\textbf{0.14} [0.10, \textbf{0.17}] &
\textbf{0.62} [\textbf{0.52}, \textbf{0.76}] &
\textbf{0.28} [\textbf{0.19}, 0.39] &
\textbf{0.52} [0.34, \textbf{0.63}] &
\textbf{0.39} [\textbf{0.29}, \textbf{0.46}] &
\textbf{26.67} & \textbf{60.00} \\
\bottomrule
\end{tabular}
\caption{Diversity and quality comparison on AIME-30 with $K{=}10$ samples per prompt. Diversity entries report median (p50) with p10/p90 in brackets; higher is better. Bold marks the better \emph{median} between methods; bracketed ranges are included for robustness context.}
\label{tab:aime30_div_quality_compact}
\end{table}
\paragraph{Diversity--quality comparison on AIME-30.}
Table~\ref{tab:aime30_div_quality_compact} summarizes diversity and quality on AIME-30 using $K{=}10$ samples per prompt.
For diversity, we report lexical diversity (distinct-1 and distinct-4) and intra-set similarity (1--self-BLEU and 1--self-ROUGE-L; higher indicates less redundancy among the $K$ samples), together with an aggregate diversity score. Each diversity entry is reported as median (p50) with p10/p90 in brackets to reflect the distribution across prompts. For quality, we report pass@1 and pass@10 based on exact-answer verification.

\paragraph{Diversity improvements are consistent across prompts.}
Compared to the GRPO baseline, Ours achieves higher median distinct-1 and distinct-4 (Table~\ref{tab:aime30_div_quality_compact}), indicating richer variation at both token and phrase granularity.
The improvements are also reflected in the bracketed ranges: p10 and p90 for distinct-4 shift upward under Ours, suggesting that the gains are not confined to a small subset of prompts but appear broadly across the evaluation set.

\paragraph{Reduced redundancy within sampled sets.}
Ours improves the median 1--self-BLEU and 1--self-ROUGE-L (Table~\ref{tab:aime30_div_quality_compact}), indicating that the $K{=}10$ sampled responses are, on average, less similar to one another.
While some tail values are comparable between methods (e.g., p90 for 1--self-ROUGE-L is similar after rounding), the median improvement supports that the typical prompt benefits from reduced intra-set redundancy.

\paragraph{Quality does not degrade with increased diversity.}
Crucially, the diversity gains are accompanied by higher pass@1 and pass@10 (Table~\ref{tab:aime30_div_quality_compact}). This suggests that increased diversity here is not merely stylistic variation but can improve the chance that at least one candidate solution is correct, improving the overall diversity--quality trade-off on AIME-30.
\subsection{Training Dynamic}
\label{appendixtraingdyanmic}
\paragraph{Validation accuracy during training.}

\begin{figure}[hbtp]
  \centering
  \begin{subfigure}[t]{0.32\linewidth}
    \centering
    \includegraphics[width=\linewidth]{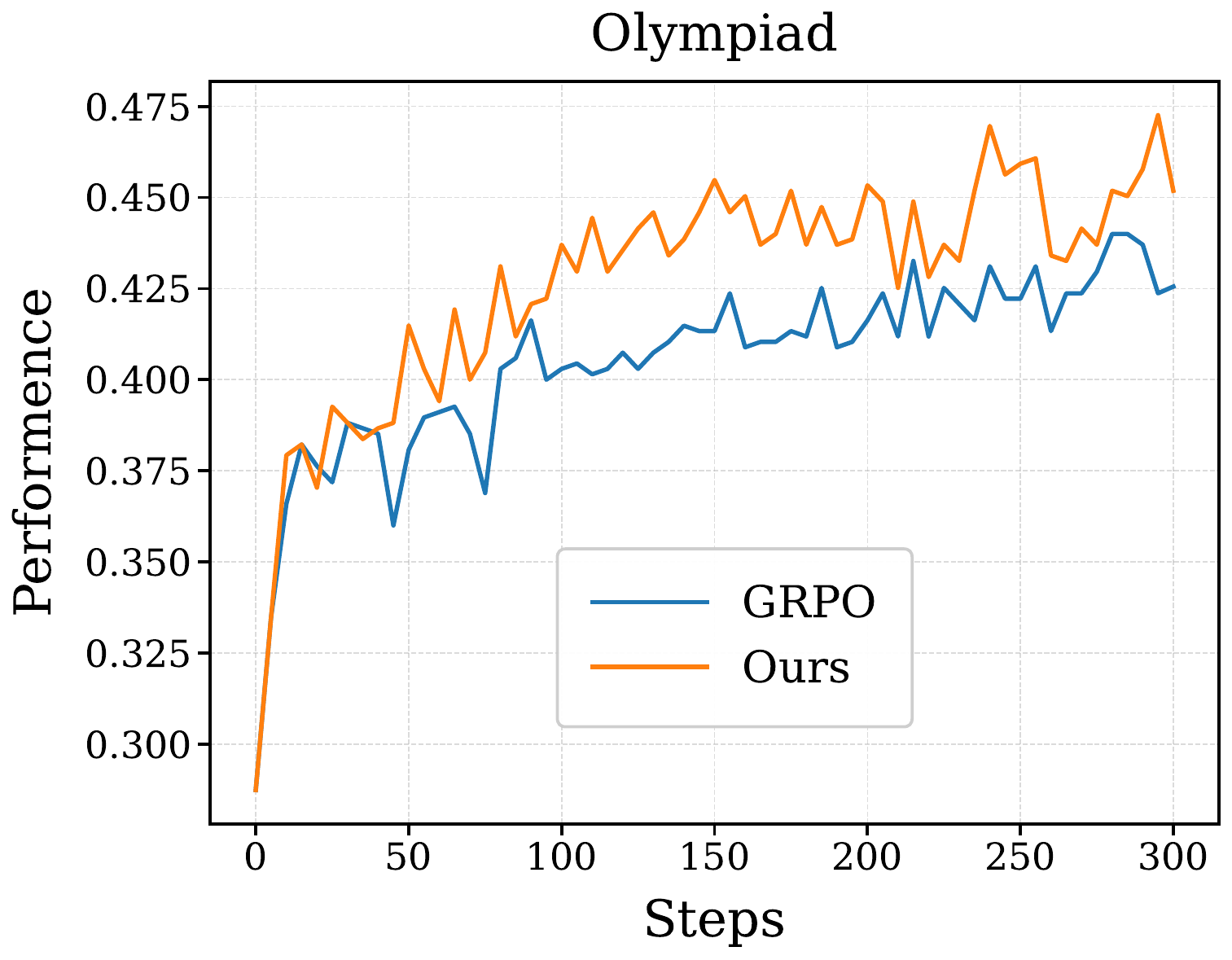}
    \caption{Olympiad}
    \label{fig:a}
  \end{subfigure}\hfill
  \begin{subfigure}[t]{0.32\linewidth}
    \centering
    \includegraphics[width=\linewidth]{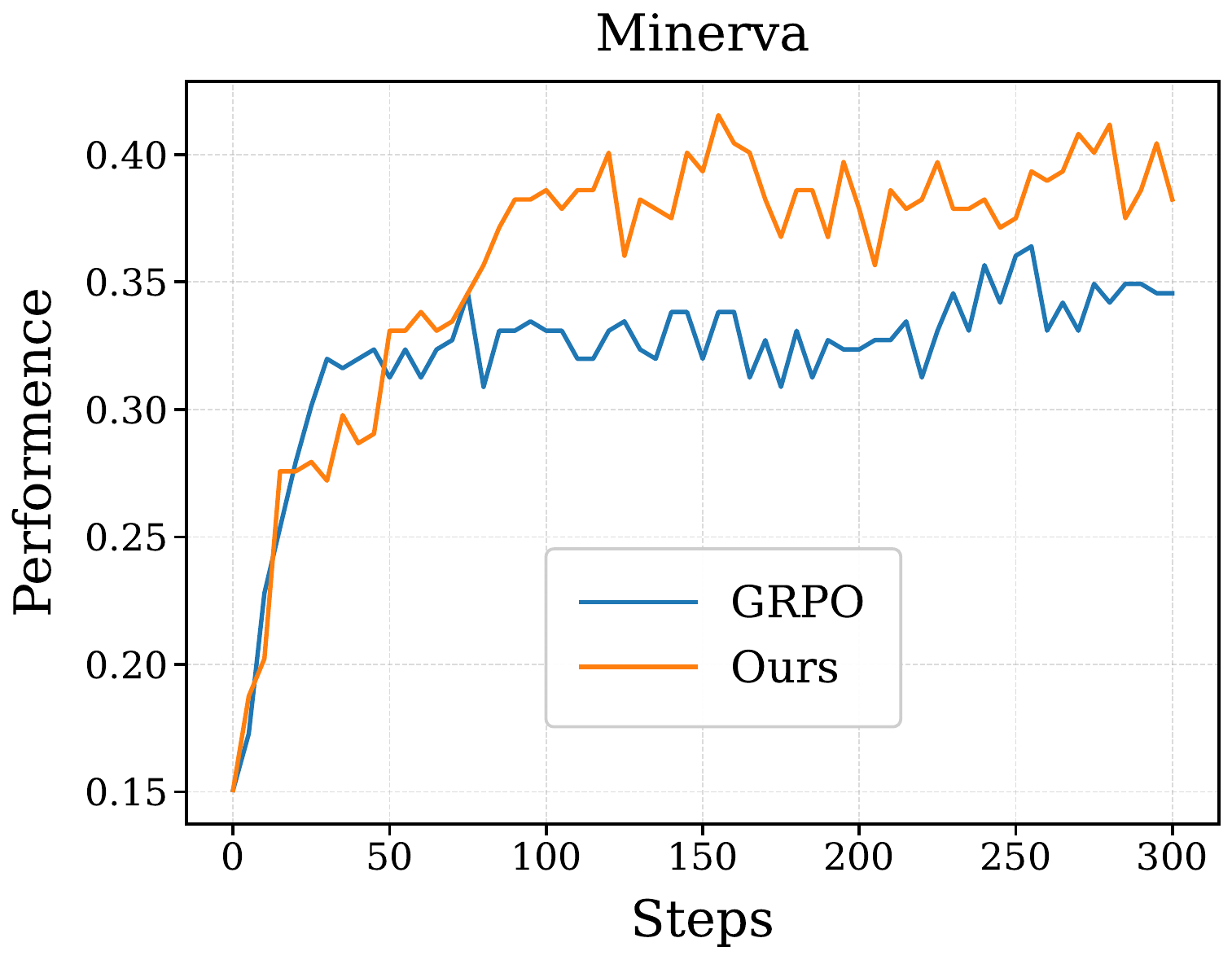}
    \caption{Minerva}
    \label{fig:b}
  \end{subfigure}\hfill
  \begin{subfigure}[t]{0.32\linewidth}
    \centering
    \includegraphics[width=\linewidth]{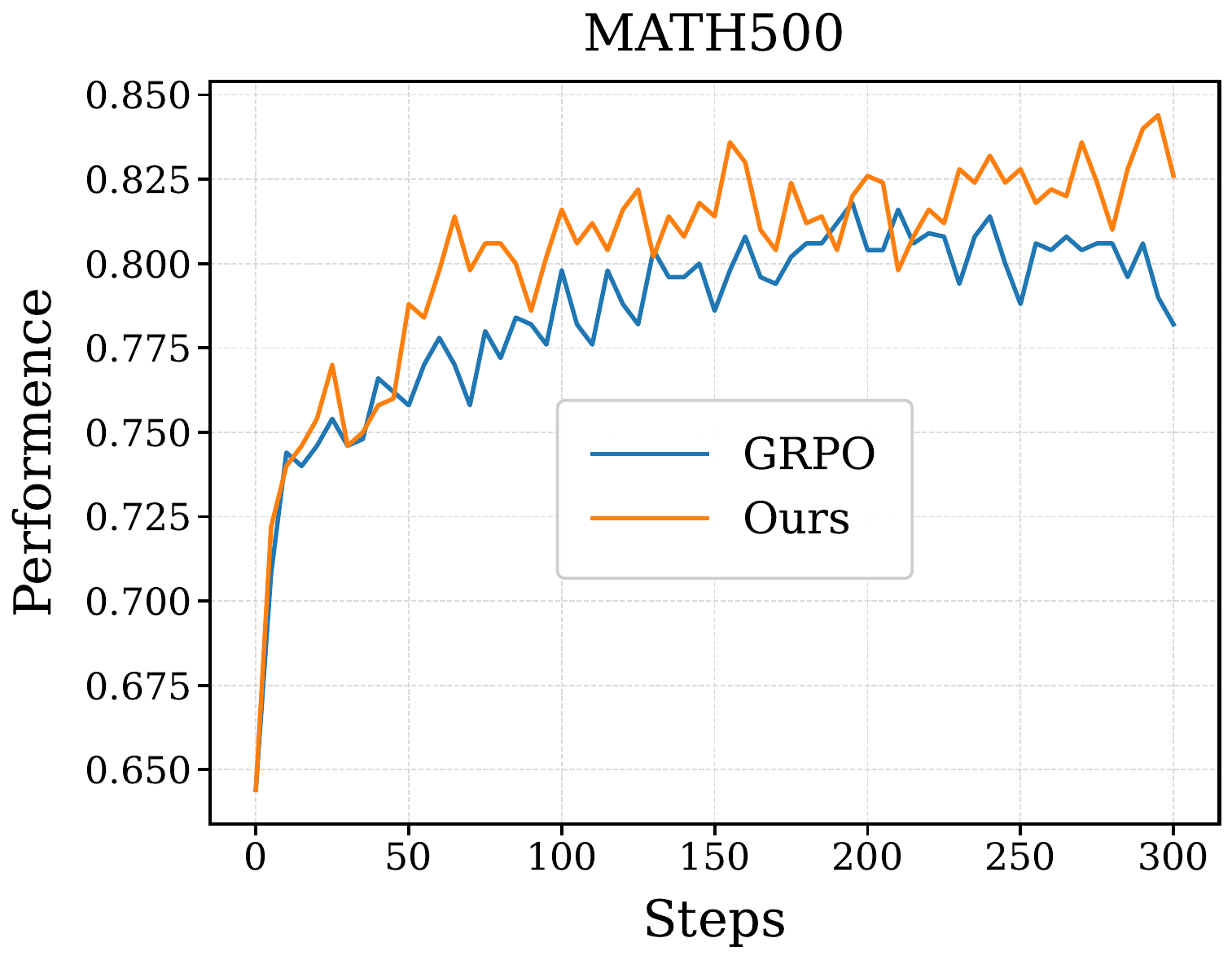}
    \caption{MATH500}
    \label{fig:c}
  \end{subfigure}
  \caption{Validation performance over training steps. We periodically evaluate the model on held-out validation sets during training. Each panel reports the performance trajectory on a different benchmark—(a) Olympiad, (b) Minerva, and (c) MATH500—comparing the GRPO baseline with our method.}
  \label{fig:val_curves}
\end{figure}

Figure~\ref{fig:val_curves} plots validation accuracy as a function of GRPO training steps on three math benchmarks. Across all datasets, our method exhibits a consistent accuracy advantage over the GRPO baseline throughout training. On Olympiad and Minerva (Figs.~\ref{fig:val_curves}a--b), the gap emerges early and remains persistent, suggesting that our approach provides more informative learning signals and avoids overly penalizing trajectories that are largely correct but fail at a late step. On MATH500 (Fig.~\ref{fig:val_curves}c), both methods improve rapidly in the initial stage, while our method maintains a stable lead in the mid-to-late stage, indicating more reliable convergence behavior. Overall, the curves in Fig.~\ref{fig:val_curves} support that salvaging partially correct rollouts translates into higher validation accuracy during training.

\paragraph{Response length behavior.}
\begin{figure}[hbtp]
  \centering
  \includegraphics[width=0.5\linewidth]{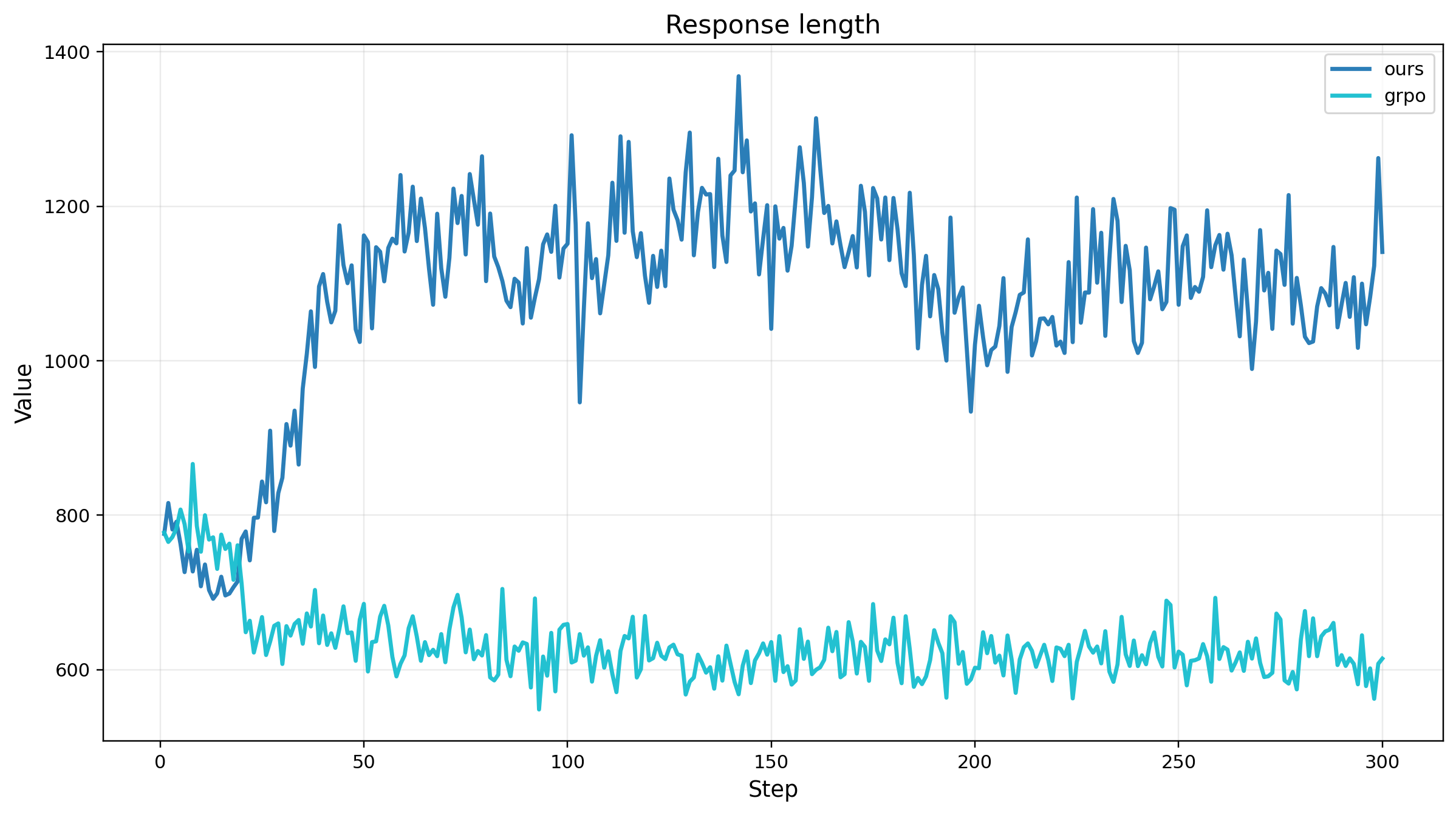}
  \caption{\textbf{Response length over training.}
  Average generated response length (in tokens) as a function of training steps, comparing our method with the GRPO baseline.}
  \label{fig:resp_len}
\end{figure}

Figure~\ref{fig:resp_len} shows the evolution of the average response length during training. Compared with GRPO, our method consistently produces substantially longer responses across most training steps. The length increases rapidly in the early stage and then stabilizes at a high level, while GRPO remains relatively short and stable throughout. This pattern is consistent with our objective of preserving and leveraging partially correct reasoning trajectories under binary outcome rewards: by avoiding overly penalizing correct prefixes and repairing only the erroneous suffix, the model is encouraged to maintain multi-step reasoning rather than collapsing to short, risk-averse outputs. Overall, Fig.~\ref{fig:resp_len} indicates that our approach sustains long-form reasoning behaviors during GRPO training.

\subsection{Reducing Discarded Near-miss Reasoning Trajectories}
\label{appendixreducenermiss}
\begin{figure}[ht]
  \centering
  \begin{subfigure}[t]{0.32\columnwidth}
    \centering
    \includegraphics[width=\linewidth]{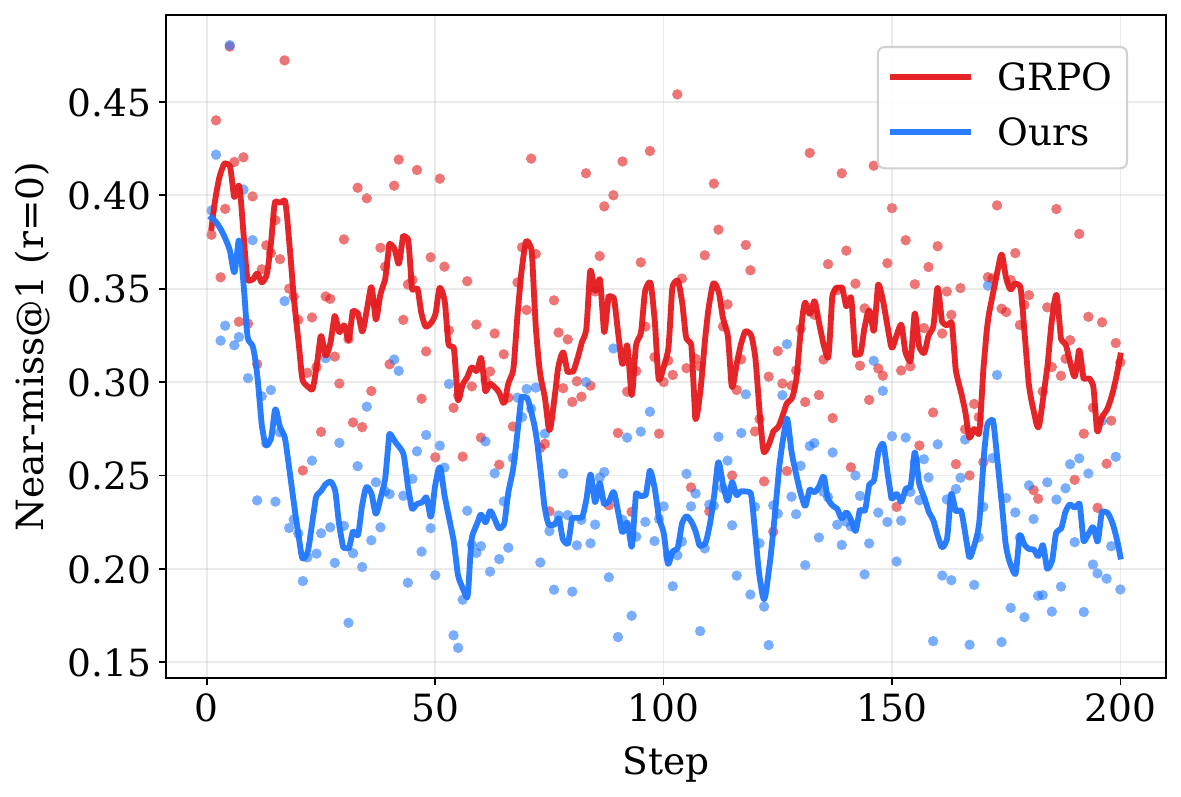}
    \caption{$k=1$}
    \label{fig:nearmiss1}
  \end{subfigure}\hfill
  \begin{subfigure}[t]{0.32\columnwidth}
    \centering
    \includegraphics[width=\linewidth]{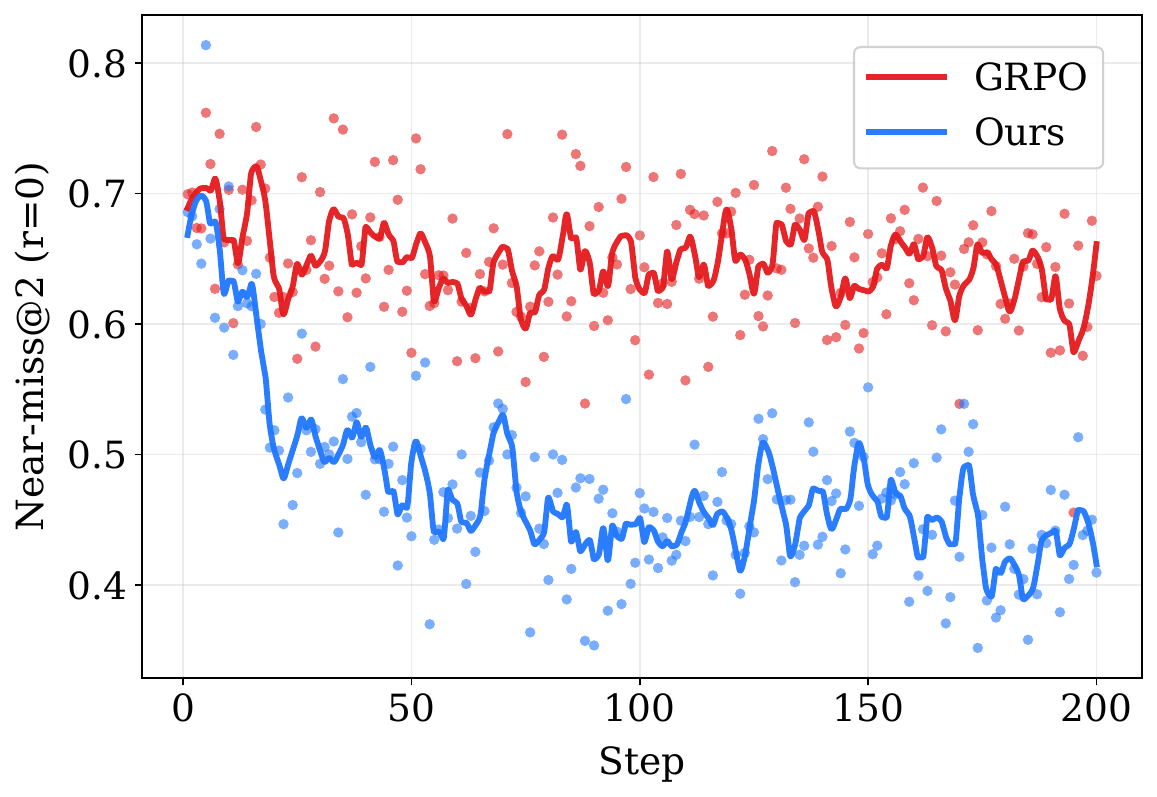}
    \caption{$k=2$}
    \label{fig:nearmiss2}
  \end{subfigure}\hfill
  \begin{subfigure}[t]{0.32\columnwidth}
    \centering
    \includegraphics[width=\linewidth]{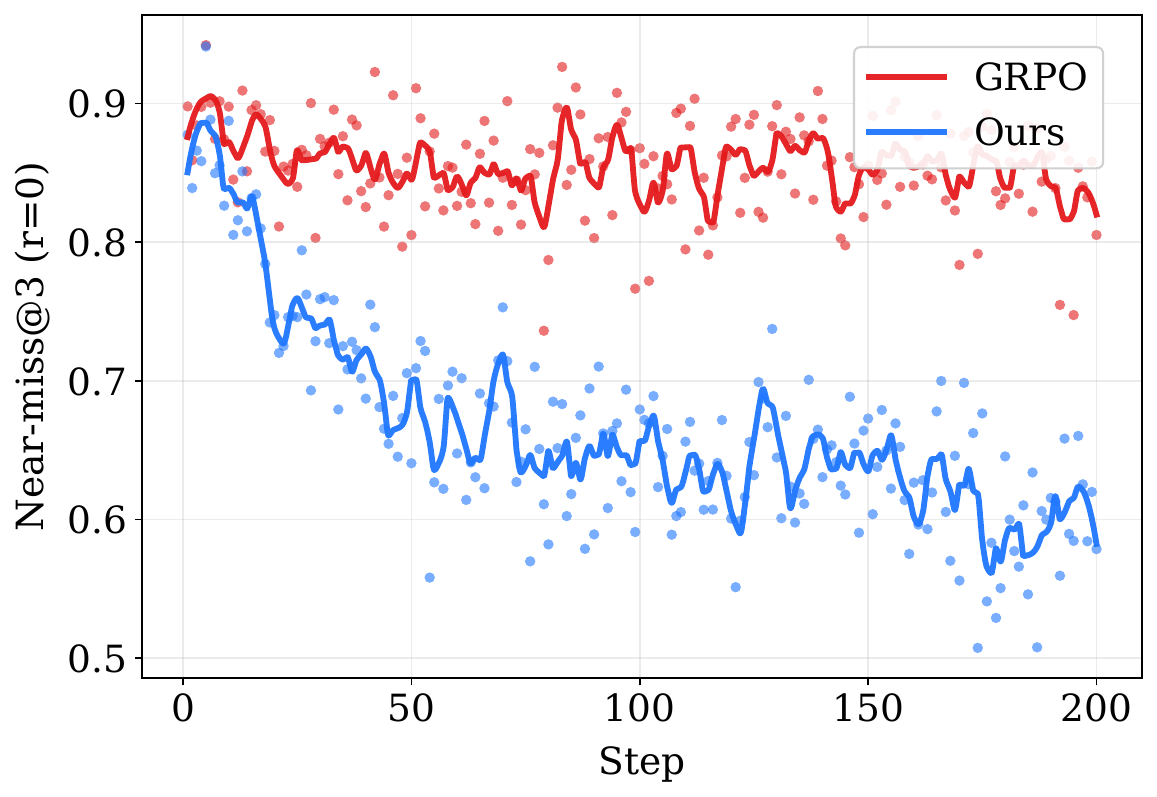}
    \caption{$k=3$}
    \label{fig:nearmiss3}
  \end{subfigure}

  \caption{\textbf{Near-miss@k (r=0) over training steps.}
  Near-miss@k denotes the fraction of zero-reward samples satisfying $N-K\le k$, where $N$ is the judged total number of steps and $K$ is the number of correct steps }
  \label{fig:nearmiss}
\end{figure}
To better quantify how much usable reasoning signal is wasted under sparse 0/1 rewards, we analyze the training dynamics of near-miss failures. We define \emph{Near-miss@k (r=0)} as the fraction of zero-reward samples satisfying $N-K \le k$, where $N$ is the judged total number of steps and $K$ is the number of correct steps (equivalently, the length of the longest correct prefix). Intuitively, Near-miss@k focuses on ``almost correct'' failures: trajectories that still receive $r=0$ but contain a largely correct chain of reasoning and deviate only in the last few steps. Under binary rewards, such trajectories are often treated as undifferentiated failures, which can discard substantial intermediate reasoning information that is potentially recoverable.

Fig.~\ref{fig:nearmiss} reports Near-miss@k (r=0) over training steps for $k\in\{1,2,3\}$. Across the entire training horizon, our method consistently yields lower Near-miss@k than the GRPO baseline, including under the strictest criterion ($k=1$). Because the metric is conditioned on $r=0$, this reduction cannot be attributed solely to a higher success rate; rather, it indicates a systematic shift in the composition of failures. In particular, fewer failures remain in the ``near-miss'' regime where a long correct prefix exists but the final outcome is blocked by a small number of late-step errors. This behavior is consistent with our goal of \emph{recycling} useful learning signal from near-miss reasoning trajectories: instead of discarding them as pure negative samples, the training process better preserves and leverages partially correct reasoning structure, which in turn improves the effectiveness of exploration.

Importantly, Near-miss@k complements standard success-rate curves by providing a within-failure diagnostic. Even when two methods exhibit comparable success rates at certain stages, differences in Near-miss@k reveal whether the remaining failures are ``close'' (suggesting recoverable structure) or ``far'' (suggesting fundamentally incorrect reasoning). The observed gap in Near-miss@k thus supports the claim that our approach reduces wasted intermediate reasoning signal in zero-reward samples and expands the effective exploration region by making more partially correct trajectories usable for subsequent refinement.
\section{Experimental Details}
\subsection{Training Details}
\label{training-details}
\paragraph{Training and infrastructure details.}
We train the model from scratch using GRPO. We conduct GRPO training for 10 epochs with a learning rate of $1\times 10^{-6}$ and a batch size of 128. For each prompt, we sample 8 rollouts per update with temperature=1. We set both the maximum prompt length and the maximum response length to 2048 tokens. Experiments are performed on \textsc{Llama}~3.1~8B, \textsc{Qwen}2.5~1.5B, and \textsc{Qwen}2.5-Math~7B. For SFT initialization, the SFT model is trained for 3 epochs(For fairness, we use DeepSeek R1 to generate the SFT data, using the same prompting template as in our GRPO sampling procedure.). All runs are implemented with the \textsc{Verl}~\cite{sheng2025hybridflow} training framework and use vLLM~\cite{kwon2023efficient} for efficient generation. All experiments are executed on a single node with 8$\times$ NVIDIA A100 GPUs.
\paragraph{Reward function.}
We adopt a binary verifiable reward based solely on the final outcome:
\begin{equation}
r(x,y)=
\begin{cases}
1, & \text{if the final answer in } y \text{ is correct under the task verifier},\\
0, & \text{otherwise.}
\end{cases}
\end{equation}
\paragraph{Rollout prompting template.}
During GRPO training, we use a unified chat-format prompting template to sample rollouts for \emph{all} models; the exact formatting is shown in Fig.~\ref{fig:rollout_template}. Concretely, each example is constructed with a system message (\texttt{You are a helpful assistant.}) followed by a user message that includes the input and an instruction to reason step by step and place the final answer within \texttt{\textbackslash boxed\{\}}. We apply this template consistently across all models and all GRPO experiments.

\paragraph{Prompt for Refiner.}
For suffix rectification, we query a strong external model (DeepSeek R1) with a continuation-only instruction, shown in Fig.~\ref{fig:r1_suffix_prompt}. The prompt provides the original problem, a verified-correct prefix (included only as context), and the ground-truth answer string, and strictly requires the model to output \emph{only} the continuation starting immediately after the prefix and to terminate with a single exact-match \texttt{\textbackslash boxed\{\}} answer. This design ensures the preserved prefix remains unchanged, while the generated suffix can be cleanly treated as off-policy supervision in our GRPO pipeline.

{ 
\centering

\begin{tcolorbox}[
  enhanced,
  breakable,       
  colback=promptbg, 
  colframe=promptbg, 
  boxrule=0pt,       
  arc=0pt,           
  boxsep=0pt,        
  left=6pt, right=6pt, top=6pt, bottom=6pt,
]
  \scriptsize\ttfamily\raggedright

  You are a meticulous math solution editor.\par
  \par
  You will be given:\par
  - PROBLEM\par
  - VERIFIED-CORRECT PREFIX (context only; do NOT repeat it)\par
  - GROUND-TRUTH final answer (string)\par
  \par
  Goal:\par
  Continue from the end of the prefix and write ONLY the continuation (the suffix).\par
  The prefix is already verified correct and will be prepended by the caller.\par
  \par
  Hard constraints:\par
  1) Output ONLY the continuation text (suffix). Do NOT output the prefix again.\par
  2) Do NOT add any preamble, headings, or meta commentary.\par
  3) Ensure every new step you write is correct.\par
  4) End with exactly one final boxed answer: \textbackslash boxed\{...\}\par
  5) The boxed answer MUST match the ground-truth answer string exactly (character-for-character).\par
  6) Do not change the problem statement; do not introduce new assumptions.\par
  \par
  Formatting guidance:\par
  - Write clear step-by-step reasoning, but keep it concise.\par
  - If you need to fix a mistake, do so only in the continuation (since the prefix is frozen).\par
  - Avoid alternative final answers; provide a single final \textbackslash boxed\{...\}.\par
  \par
  PROBLEM:\par
  \{problem\_text\}\par
  \par
  GROUND-TRUTH ANSWER (must match exactly):\par
  \{ground\_truth\}\par
  \par
  VERIFIED-CORRECT PREFIX (for context only; do not repeat):\par
  \textless\textless\textless PREFIX\_START\par
  \{prefix\_text\}\par
  PREFIX\_END\textgreater\textgreater\textgreater\par
  \par
  Write ONLY the continuation starting immediately after the prefix.\par
\end{tcolorbox}

\captionof{figure}{\textbf{Continuation prompt for suffix rectification.}
Prompt used to query DeepSeek R1: given the problem, a verified-correct prefix, and the ground-truth answer string, the model outputs only the continuation (suffix) and ends with an exact-match \texttt{\textbackslash boxed\{\}} answer.}
\label{fig:r1_suffix_prompt}

} 
{ 
\centering

\begin{tcolorbox}[
  enhanced,
  breakable,       
  colback=promptbg, 
  colframe=promptbg, 
  boxrule=0pt,       
  arc=0pt,           
  boxsep=0pt,        
  left=6pt, right=6pt, top=6pt, bottom=6pt,
]
  \scriptsize\ttfamily\raggedright
  <|im\_start|>system\par
  You are a helpful assistant.<|im\_end|>\par
  <|im\_start|>user\par
  \{input\}\par
  Please reason step by step, and put your final answer within \textbackslash boxed\{\}.<|im\_end|>\par
  <|im\_start|>assistant\par
  \{output\}\par
\end{tcolorbox}

\captionof{figure}{\textbf{Rollout prompting template.} Chat-format prompt used to sample rollouts during GRPO training.}
\label{fig:rollout_template}

} 

\subsection{PRM Utilization}
\label{prm-details}

We employ \textsc{Qwen2.5-Math-PRM-7B}~\cite{zhang2025lessons} as our step-level verifier, adhering to the official open-source configuration. To perform step-wise scoring, we first segment the raw model solution into individual reasoning steps using the double newline delimiter (\texttt{\textbackslash n\textbackslash n}), as observed in standard chain-of-thought outputs. These steps are then concatenated with the special token \texttt{<extra\_0>} to construct the prompt. The model calculates the probability of the positive class for the token mask corresponding to these separators, which we extract as the step-level reward. We establish a validity threshold of $\tau = 0.5$, classifying steps with rewards strictly less than $0.5$ as incorrect, while those with rewards greater than or equal to $0.5$ are deemed correct.
{ 
\centering
\definecolor{codebg}{gray}{0.95}

\begin{tcolorbox}[
  enhanced,
  breakable,       
  colback=codebg,
  colframe=codebg,
  boxrule=0pt,
  arc=0pt,
  boxsep=0pt,
  left=6pt, right=6pt, top=6pt, bottom=6pt,
]
  \scriptsize\ttfamily
\begin{lstlisting}[
    language=Python, 
    breaklines=true, 
    basicstyle=\scriptsize\ttfamily, 
    columns=fullflexible, 
    showstringspaces=false
]
model_name = "Qwen/Qwen2.5-Math-PRM-7B"

# Sample input: The response list is obtained by splitting the raw output by '\n\n'
data = {
    "system": "Please reason step by step, and put your final answer within \\boxed{}.",
    "query": "Sue lives in a fun neighborhood.  One weekend, the neighbors decided to play a prank on Sue... (omitted for brevity) ... At noon on Sunday, how many more pink plastic flamingos were out than white plastic flamingos?",
    "response": [
      "To find out how many more pink plastic flamingos were out than white plastic flamingos at noon on Sunday, we can break down the problem into steps. First, on Friday, the neighbors start with 18 pink plastic flamingos.",
      "On Saturday, they take back one third of the flamingos. Since there were 18 flamingos, (1/3 \\times 18 = 6) flamingos are taken back. So, they have (18 - 6 = 12) flamingos left in their possession. Then, they paint these 6 flamingos white and put them back out on Sue's front yard. Now, Sue has the original 12 pink flamingos plus the 6 new white ones. Thus, by the end of Saturday, Sue has (12 + 6 = 18) pink flamingos and 6 white flamingos.",
      "On Sunday, the neighbors add another 18 pink plastic flamingos to Sue's front yard. By the end of Sunday morning, Sue has (18 + 18 = 36) pink flamingos and still 6 white flamingos.",
      "To find the difference, subtract the number of white flamingos from the number of pink flamingos: (36 - 6 = 30). Therefore, at noon on Sunday, there were 30 more pink plastic flamingos out than white plastic flamingos. The answer is (\\boxed{30})."
    ]
}

# Construct the conversation with special separator tokens
messages = [
    {"role": "system", "content": data['system']},
    {"role": "user", "content": data['query']},
    {"role": "assistant", "content": "<extra_0>".join(data['response']) + "<extra_0>"},
]

# Apply chat template and encode
conversation_str = tokenizer.apply_chat_template(
    messages, tokenize=False, add_generation_prompt=False
)
input_ids = tokenizer.encode(conversation_str, return_tensors="pt").to(device)

# Forward pass to get logits
outputs = model(input_ids=input_ids)

# Extract rewards at separator token positions
step_sep_id = tokenizer.encode("<extra_0>")[0]
token_masks = (input_ids == step_sep_id)
step_reward = make_step_rewards(outputs[0], token_masks)
# Output: [[1.0, 0.1904296875, 0.9765625, 1.0]]
\end{lstlisting}
\end{tcolorbox}

\captionof{figure}{Implementation details for extracting step-wise rewards. We first split the raw solution by \texttt{\textbackslash n\textbackslash n} to obtain the step list (as shown in \texttt{data['response']}), then join them with the \texttt{<extra\_0>} token to query the PRM.}
\label{fig:prm_code}

} 
\newcommand{\nnsep}{\colorbox{yellow!40}{\strut\texttt{\textbackslash n\textbackslash n}}}

{ 
\centering

\begin{tcolorbox}[
  enhanced,
  breakable,        
  colback=gray!6,
  colframe=black!20,
  boxrule=0.6pt,
  arc=2pt,
  left=6pt,right=6pt,top=6pt,bottom=6pt,
  title={\textbf{Base-model outputs contain explicit step boundaries via \texttt{\textbackslash n\textbackslash n}}},
  fonttitle=\bfseries,
  pad at break=2pt
]

\textbf{Observation.} We segment solutions into intermediate steps using the literal delimiter \nnsep for PRM-based step-wise scoring. The examples below show that \nnsep frequently appears in outputs from several base models, making step boundary detection straightforward in practice.

\vspace{1.5mm}

\begin{tcolorbox}[
  enhanced, 
  breakable, 
  colback=blue!3, colframe=blue!35,
  boxrule=0.6pt, arc=2pt,
  left=5pt,right=5pt,top=4pt,bottom=4pt,
  title={\textbf{Llama 3.1-8B}},
  fonttitle=\bfseries,
]
\noindent
On Monday at work, David produces $w$ widgets per hour, and works for $t$ hours. Exhausted by this work, on Tuesday, he decides to work for $2$ fewer hours, but manages to produce $4$ additional widgets per hour. If $w = 2t$, how many more widgets did David produce on Monday than on Tuesday?
\nnsep
We are given that on Monday, David produces $w$ widgets per hour and works for $t$ hours. Therefore, the total number of widgets produced on Monday is: $\text{Monday total}=w\times t$.
\nnsep
On Tuesday, he works for $t-2$ hours and produces $w+4$ widgets per hour. Thus, $\text{Tuesday total}=(w+4)(t-2)$.
\nnsep
We are also given $w=2t$. We want $\text{Difference}=wt-(w+4)(t-2)$. Expanding $(w+4)(t-2)=wt-2w+4t-8$, so $\text{Difference}=2w-4t+8$. Substituting $w=2t$ gives $\text{Difference}=8$. Therefore, the answer is $\boxed{8}$.
\end{tcolorbox}

\vspace{1.5mm}

\begin{tcolorbox}[
  enhanced, 
  colback=green!3, colframe=green!35,
  boxrule=0.6pt, arc=2pt,
  left=5pt,right=5pt,top=4pt,bottom=4pt,
  title={\textbf{Qwen2.5-1.5B}},
  fonttitle=\bfseries,
  breakable,
]
\noindent
Let $p(x)=a_n x^n+a_{n-1}x^{n-1}+\dots+a_0$ with each $a_i$ a nonnegative integer. We are given $p(1)=4$ and $p(5)=136$.
\nnsep
Since $p(1)=a_n+\dots+a_0=4$, the coefficients are small and each $a_i\le4$. Write $136$ in base 5: $136=1\cdot5^3+0\cdot5^2+2\cdot5+1=1021_5$.
\nnsep
Try degree 3: $125a_3+25a_2+5a_1+a_0=136$ with $a_3+a_2+a_1+a_0=4$. Setting $a_3=1$ yields $25a_2+5a_1+a_0=11$ and $a_2+a_1+a_0=3$, giving $6a_2+a_1=2$. Thus $a_2=0,a_1=2,a_0=1$ and $p(x)=x^3+2x+1$.
\nnsep
Then $p(6)=6^3+2\cdot6+1=229$, so $\boxed{229}$.
\end{tcolorbox}

\vspace{1.5mm}

\begin{tcolorbox}[
  enhanced, 
  colback=red!2, colframe=red!30,
  boxrule=0.6pt, arc=2pt,
  left=5pt,right=5pt,top=4pt,bottom=4pt,
  title={\textbf{Qwen2.5-Math-7B}},
  fonttitle=\bfseries,
  breakable,
]
\noindent
To find the area of the quadrilateral with vertices $(1,3)$, $(1,1)$, $(2,1)$, and $(2006,2007)$, use the Shoelace formula:
$\text{Area}=\frac12\left|x_1y_2+x_2y_3+x_3y_4+x_4y_1-(y_1x_2+y_2x_3+y_3x_4+y_4x_1)\right|$.
\nnsep
Substituting the points gives $1\cdot1+1\cdot1+2\cdot2007+2006\cdot3=10034$ and $3\cdot1+1\cdot2+1\cdot2006+2007\cdot1=4024$, so the difference is $6010$.
\nnsep
Thus $\text{Area}=\frac12\cdot6010=3005$, so the answer is $\boxed{3005}$.
\end{tcolorbox}

\end{tcolorbox}

\captionof{figure}{Examples from three base models showing that the literal delimiter \nnsep frequently appears inside the generated solution text. We highlight \nnsep inline (without introducing extra line breaks), demonstrating that it can be used to segment step boundaries for PRM-based step-wise scoring.}
\label{fig:prm_step_delim_inline}

} 

\section{SCOPE Example}
\label{Appendixexample}

\begin{figure*}[htbp]
    \centering
    \includegraphics[width=1.1\linewidth]{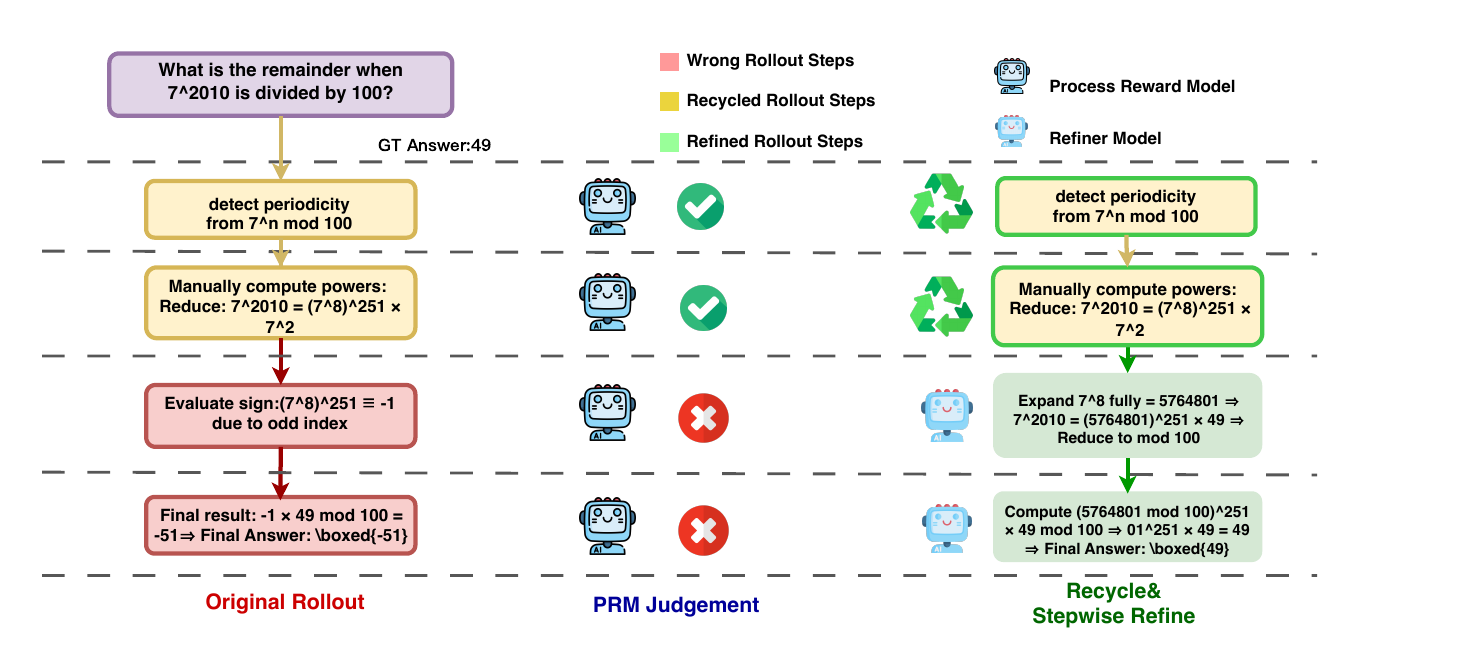}
    \caption{\textbf{SCOPE Rectification Process.} Illustrated on a specific SCOPE rollout, for a selected suboptimal trajectory, the Process Reward Model (PRM) identifies and recycles valid steps based on confidence scores and a validity threshold. The subsequent erroneous steps are then targeted by the Refiner Model, which performs \textbf{off-policy guided step-wise rectification} to correct the reasoning logic, ultimately synthesizing a fully correct rollout.}
    \label{fig:example}
\end{figure*}

\paragraph{Case Study: Step-wise Rectification.}
Figure~\ref{fig:example} provides a concrete demonstration of the SCOPE mechanism on a modular arithmetic problem ($7^{2010} \pmod{100}$). In the original rollout, the policy correctly identifies the periodicity and performs initial reductions, receiving high PRM verification scores. However, the reasoning collapses at the third step where the model hallucinates a sign rule, causing the PRM confidence to drop significantly, below the validity threshold. Rather than discarding this near-miss attempt, SCOPE \textbf{recycles} the verified prefix (Steps 1-2) and triggers the Refiner Model at the precise locus of error. The Refiner performs off-policy guided step-wise rectification to generate a valid continuation, correctly expanding the terms and computing the modulo operation. This process effectively salvages the partially correct trajectory, converting a zero-reward failure into a high-quality training signal.

\end{document}